\documentclass[letterpaper]{article} 
\usepackage[draft]{aaai2026}  
\usepackage{times}  
\usepackage{helvet}  
\usepackage{courier}  
\usepackage[hyphens]{url}  
\usepackage{graphicx} 
\usepackage{natbib}  
\usepackage{caption} 
\usepackage{algorithm}
\usepackage{algorithmic}
\newcommand{\new}[1]{\textcolor{black}{#1}}
\usepackage{newfloat}
\usepackage{listings}
\DeclareCaptionStyle{ruled}{labelfont=normalfont,labelsep=colon,strut=off} 
\floatstyle{ruled}
\newfloat{listing}{tb}{lst}{}
\floatname{listing}{Listing}
\usepackage{amsmath}
\usepackage{multirow}
\usepackage[table,xcdraw]{xcolor}
\usepackage{amssymb}
\usepackage{siunitx}
\usepackage{newfloat}
\usepackage{listings}

\usepackage{algorithm}
\usepackage{algorithmic}

\title{GEWDiff: Geometric Enhanced Wavelet-based Diffusion Model for Hyperspectral Image Super-resolution}
\author {
    Sirui Wang\textsuperscript{\rm 1},
    Jiang He\textsuperscript{\rm 1},
    Natàlia Blasco Andreo\textsuperscript{\rm 2},
    Xiao Xiang Zhu \textsuperscript{\rm 1,3}\thanks{Corresponding author.}
}
\affiliations {
    \textsuperscript{\rm 1}Technical University of Munich, Arcisstraße 21, 80333 Munich, Germany\\
    \textsuperscript{\rm 2}Universitat Autònoma de Barcelona, 08193 Cerdanyola del Vallès, Barcelona, Spain\\
    \textsuperscript{\rm 3}Munich Center for Machine Learning, 80333 Munich, Germany\\
    sirui.wang@tum.de, Natalia.Blasco@uab.cat, jiang.he@tum.de,xiaoxiang.zhu@tum.de
}

\usepackage{bibentry}

\begin{document}

\maketitle

\begin{abstract}
Improving the quality of hyperspectral images (HSIs), such as through super-resolution, is a crucial research area. However, generative modeling for HSIs presents several challenges. Due to their high spectral dimensionality, HSIs are too memory-intensive for direct input into conventional diffusion models. Furthermore, general generative models lack an understanding of the topological and geometric structures of ground objects in remote sensing imagery. In addition, most diffusion models optimize loss functions at the noise level, leading to a non-intuitive convergence behavior and suboptimal generation quality for complex data. To address these challenges, we propose a Geometric Enhanced Wavelet-based Diffusion Model (GEWDiff), a novel framework for reconstructing hyperspectral images at 4-times super-resolution. A wavelet-based encoder-decoder is introduced that efficiently compresses HSIs into a latent space while preserving spectral-spatial information. To avoid distortion during generation, we incorporate a geometry-enhanced diffusion process that preserves the geometric features. Furthermore, a multi-level loss function was designed to guide the diffusion process, promoting stable convergence and improved reconstruction fidelity. Our model demonstrated state-of-the-art results across multiple dimensions, including fidelity, spectral accuracy, visual realism, and clarity.
\end{abstract}

%
 \begin{links}
     \link{Code}{https://github.com/zhu-xlab/GEWDiff}
 \end{links}

\section{Introduction}
\label{sec:intro}
Hyperspectral images (HSIs) offer a unique perspective by capturing continuous spectral features of ground objects. Despite advancements in research,  the high costs and low coverage of super-resolution (SR) hyperspectral data limit their applications. Currently, open-access hyperspectral airborne data are predominantly regional, typically focused on several cities, increasing their exclusivity and cost. Hyperspectral satellites have better coverage but suffer from insufficient spatial resolution. Improving the spatial resolution of hyperspectral satellite images is, therefore, crucial to allow for fully harnessing the potential of hyperspectral data in Earth observation (EO). Many fusion models that combine hyperspectral and multispectral images (MSIs) have been developed. However, the fusion model cannot generate HSIs in any region of interest without any prior knowledge offered by VHR RGB data. Most fusion models can obtain 10 m resolution hyperspectral data with MSIs, such as Sentinel 2, but to get a spatial resolution from 10 to 2.5 m, we focused on the single-image-generation methods for HSIs at 4-times super-resolution. 
\begin{figure*}[!ht]
  \centering
  \includegraphics[width = 0.85\linewidth]{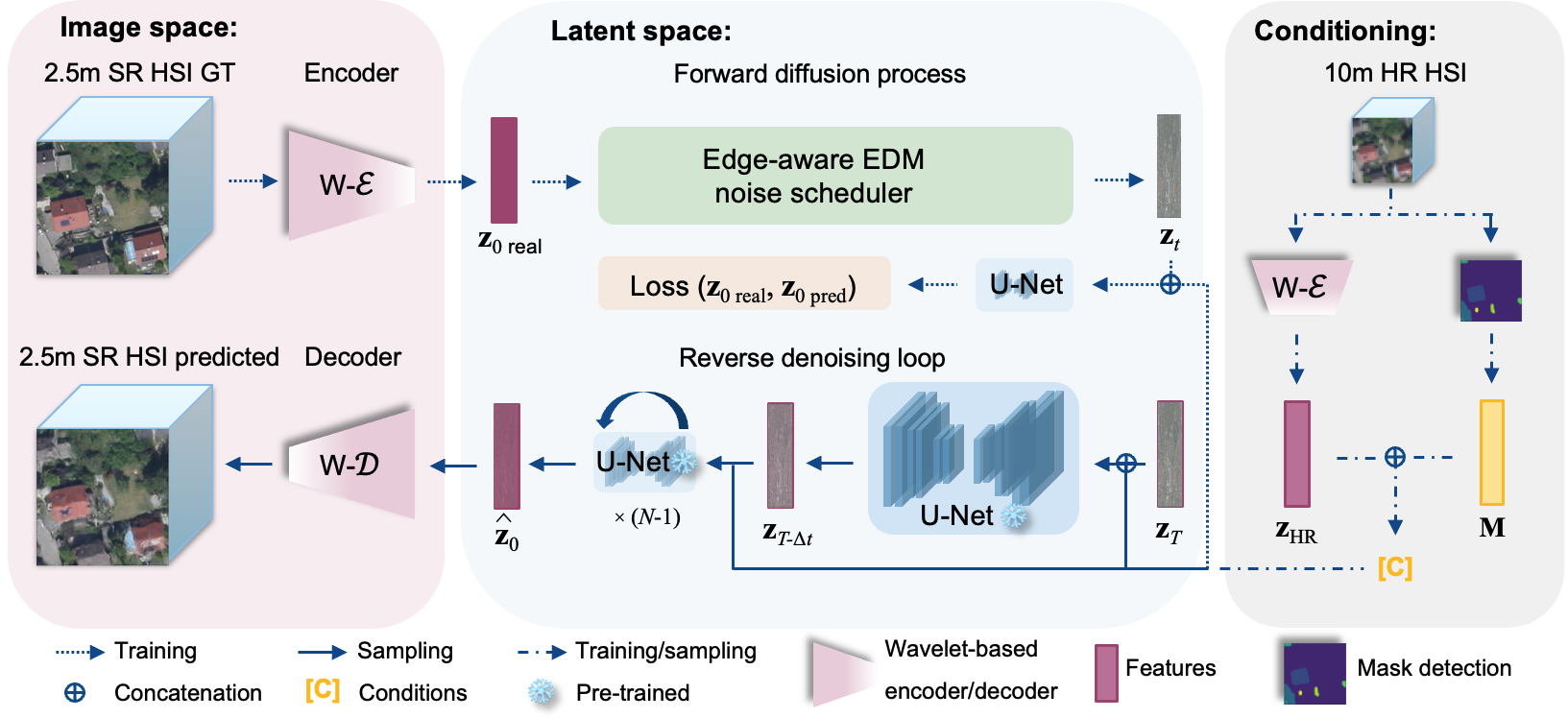}
  \caption{\new{Illustration of the Geometric Enhanced Wavelet-based Diffusion Model pipeline.}}
  \label{fig:pipe}
\end{figure*}

Traditional hyperspectral image (HSI) super-resolution approaches typically rely on interpolation techniques, such as nearest neighbor or bilinear interpolation. While straightforward and computationally efficient, these methods fail to capture the complex, nonlinear relationships present in high-dimensional spectral data. Recently, deep learning-based methods, such as convolutional neural networks (CNNs) and generative adversarial networks (GANs) \cite{sidorov2019deep,mcnet,hou2022deep,legan,bagan}, have emerged as powerful alternatives. These models are capable of learning expressive spectral-spatial representations directly from data, leading to significant improvements in reconstruction accuracy over classical methods. Recent studies have shown the strong potential of transformers \cite{zhang2023essaformer,chen2023msdformer,yu2023dstrans,{su2025eigensr}} in vision tasks, due to their ability to model long-range spatial and spectral dependencies. However, a common limitation shared by the aforementioned methods is their difficulty in generating rich textures and complex spatial structures, despite their strong performance in preserving spectral fidelity. 

Diffusion models have recently demonstrated remarkable success in generating high-quality natural images, as seen in models such as the Stable Diffusion \cite{rombach2021stablediff} and DiT \cite{peebles2023scalablediffusionmodelstransformers} frameworks. Motivated by their strong generative capabilities and robustness in modeling complex distributions, researchers have begun exploring their applicability to hyperspectral image super-resolution. For instance, SpectralDiff \cite{chen2023spectraldiff} and HSR-Diff \cite{wu2023hsrdiff} extend the diffusion paradigm directly to the hyperspectral domain, aiming to better model spectral-spatial correlations. However, despite recent progress, adapting diffusion models to hyperspectral image generation remains a significant challenge. Unlike natural or multispectral images, HSI data typically exhibit a much lower signal-to-noise ratio and higher spectral dimensionality, making it difficult to design diffusion architectures that balance spatial fidelity, spectral accuracy, and visual realism. Existing architectures often suffer from a slow model convergence speed, excessive sampling steps, and a need for high computational GPU memory, which limits their practicality in real-world scenarios. We propose a Geometric Enhanced Wavelet-based Diffusion Model (GEWDiff) to address these challenges, as shown in Figure \ref{fig:pipe}. The main contributions of this study are summarized as follows:
\begin{itemize}
    \item \textbf{An efficient wavelet-based encoder-decoder} will almost losslessly transform hyperspectral data into a latent space by decomposing the input into multiple frequency levels. This method preserves spectral-spatial information while reducing channel dimensionality without long-term training. 
    \item \textbf{Structural invariance control via the diffusion process:} An edge-aware noise scheduler was designed to improve generation efficiency and accuracy during training. Mask conditioning ensures preservation of the geometric integrity and prevents distortion.
    \item \textbf{A multi-level loss function comprising pixel-wise loss, perceptual loss, and gradient loss:} Each part of the loss function contributes to a balanced convergence speed. The loss function also enables the evaluation and alignment of predicted and ground-truth images on specific details and semantic understanding.
\end{itemize}

\section{Related Work}
\subsection{Properties of wavelet-based diffusion models}
\label{sec:wavetrw}
Wavelet-based diffusion models have recently gained increasing attention due to their suitability for image generation. For example, WaveDiff \cite{phung2023wavelet} demonstrated that applying diffusion in the wavelet domain allows images to be compressed into a structured latent space, enabling almost-lossless reconstruction while significantly reducing computational overheads. This approach not only preserves high-fidelity details but also offers substantial memory savings, making it especially advantageous for large-scale diffusion models. Building on this, Zhao et al. \cite{Zhao_2024_underwater} proposed a parallel diffusion strategy that separates high-frequency and low-frequency components for underwater image restoration. Shi et al. \cite{shi2024wavediffurdiffusionsdebasedsolver} proposed WaveDiffUR, a wavelet-domain diffusion model for remote sensing ultra-resolution (UR), which involved reformulating high-magnification SR as a conditional stochastic differential equation (SDE) solved via iterative wavelet decomposition, integrating pre-trained SR modules for scalability and a cross-scale pyramid (CSP) constraint to preserve spectral-spatial fidelity. Si et al. \cite{hsireconwavelet} proposed CASSIDiff, the first diffusion model for CASSI hyperspectral reconstruction, which integrated a DWT-based feature fusion mechanism to reduce noise and a spectral-spatial attention module to capture spectral correlations. Despite the success of various wavelet-based diffusion models in natural and remote sensing images, their application to hyperspectral image generation remains unexplored.

\subsection{Diffusion model for hyperspectral image super-resolution}
\label{sec:diffrw}
Recent work has proposed adapting the diffusion process to better fit the characteristics of HSI data. As such, existing approaches can be broadly categorized into three paradigms: two-stage models, grouped autoencoder models, and end-to-end frameworks \cite{wang2023hyperspectralreview}. Two-stage models decompose the super-resolution task into two separate subtasks handled by distinct networks. For example, HSI-Gene \cite{pang2024hsigenefoundationmodelhyperspectral} first generates high-resolution RGB bands from the input HSI, and then fuses them with the low-resolution HSI to reconstruct the final output. This modular design helps reduce computational complexity and allows for flexible training. Grouped models partition the spectral bands into groups and process them in parallel, often using autoencoder-style architectures. DMGASR \cite{wang2024dmgasr}, for instance, employs spectral grouping and trains separate VAE-based diffusion modules for different groups, enabling scalable training across high-dimensional spectra while preserving inter-band correlations. End-to-end frameworks perform full-spectrum reconstruction in a single model, often incorporating various strategies to manage complexity and improve the expressiveness. For example, HIR-Diff \cite{pang2024hirdiff} integrates a codebook with singular value decomposition to compress and guide the generation process. MTLSC-Diff \cite{qu2024mtlscdiff} uses classification maps as spatial priors to improve the generation accuracy. LSDiff \cite{cheng2024lsdiff} applies the diffusion process in a compressed latent space, which allows reducing memory usage while maintaining the generation quality. \new{Although some methods have been explored, most rely on two-stage training or fail to simultaneously ensure spectral fidelity and visual quality, while our model addresses both challenges effectively.}

\section{Method}
\subsection{Wavelet-based encoder and decoder}
\label{sec:ende}
Our encoder-decoder is based on Regression wavelet analysis (RWA), first proposed for lossless hyperspectral image compression by \cite{rwt}. RWA applies a predefined number of Haar wavelet \cite{haar1910orthogonalen} decompositions intercalated with a linear regression of the spectral dimension to exploit the redundancy that still remains in the discrete wavelet transform (DWT) domain to further compress the data. RWA can compress HSIs with a more efficient, lossless, or near-lossless transform by storing the prediction error. The structure is shown in Figure \ref{fig:ende}.

\subsubsection{Encoder.} For the super-resolution task, RWA allows us to reduce the number of bands given to the diffusion model by using the Haar wavelet. Let $\textbf{I}_\textrm{LR}$ be the input 10 m high-resolution hyperspectral image, the J-th level RWA transform can be represented as:
\begin{equation}
    (\textbf{V}_{\textrm{LR}}^{J},(\textbf{w}_{\textrm{LR}}^{j})^{1\leqslant j\leqslant J})=\textrm{RWA}(\textbf{I}_{\textrm{LR}},J),
\end{equation}
\begin{equation}\label{eq:rwa_enc}
\hat{\textbf{w}}_i^j = \beta_{i,0}^j + \beta_{i,1}^j \textbf{V}_1^j + ... +  \beta_{i,k}^j \textbf{V}_k^j ,
\end{equation}
\begin{equation}\label{eq:rwa_minimization}
\textrm{min} ||\textbf{w}_i^j - \hat{\textbf{w}}_i^j||_2,
\end{equation}
where $\textbf{w}_i^j$ represents the i-th details (high-coefficient) of the j-th level wavelet transform and $\hat{\textbf{w}}_i^j$ its prediction; $\textbf{V}_k^j$ represents the k-th low-coefficient (main-coefficient) of the j-th level wavelet transform and $\beta_{i,k}^j$ the linear regression coefficients that will be learned to adjust the linear regression. Contrary to traditional RWA, where the residuals 
\begin{equation}\label{eq:rwa_residual}
\textbf{W}_{\textrm{LR}}^j = \textbf{w}_{\textrm{LR}}^j - \hat{\textbf{w}}_{\textrm{LR}}^j,
\end{equation}
are computed in order to fully recover the original signal, the proposed encoder will only store $\textbf{V}_{\textrm{LR}}^J$ and the weights of all the adjusted linear models $\textbf{B}_{\textrm{LR}} = [\beta^{1\leqslant j\leqslant J}]$, where $J$ is the level of wavelet transforms applied. The main coefficients $\textbf{V}_{\textrm{LR}}^{J}$, which contain the most critical information, are used as input for the principal component analysis (PCA). The following PCA transformation enables a more efficient compression of hyperspectral imagery (HSI) by achieving a higher compaction factor while preserving more information. Furthermore, it can convert the sparse wavelet-based coefficients into a dense and orthogonal matrix, facilitating more coherent spectral analysis:
\begin{equation}
    (\textbf{z}_{\textrm{LR}}, \textbf{R}_{\textrm{LR}}) = \textrm{PCA} (\textbf{V}_{\textrm{LR}}^{J}),
\end{equation}
where $\textbf{z}_{\textrm{LR}}$ is the feature that will be input to the diffusion latent space, and $\textbf{R}_{\textrm{LR}}$ represents the remaining components that will be kept and reused in the decoder equation (\ref{eq:6}).
\subsubsection{Decoder.} For the reconstruction, the inverse PCA recovers predicted features $\hat{\textbf{z}}_{0}$ from the diffusion process to the super-resolution HSI main coefficients $\hat{\textbf{V}}_{\textrm{SR}}^{J} $.
\begin{equation}
    (\hat{\textbf{V}}_{\textrm{SR}}^{J} )= \textrm{I-PCA} (\hat{\textbf{z}}_{0},\textbf{R}_{\textrm{LR}}).
    \label{eq:6}
\end{equation}
The final super-resolved image is obtained by an inverse RWA, having set the residuals $\textbf{W}_{\textrm{LR}}^j$ to zero, since these are not available for the super-resolution HSI image:
\begin{equation}\label{eq:rwa_recon}
(\hat{\textbf{I}}_{\textrm{SR}})= \textrm{I-RWA}(\hat{\textbf{V}}_{\textrm{SR}}^{J},\textbf{B}_\textrm{LR}, \textbf{W}_{\textrm{LR}}^j, J).
\end{equation}
The details $\hat{\textbf{w}}_{\textrm{SR}}^j$ will be predicted by the adjusted linear regression model $\textbf{B}_\textrm{LR}$ to recover the information lost in the wavelet transform. Once the diffusion outputs the super-resolution components $\hat{\textbf{V}}_{\textrm{SR}}^{J}$, inverse-RWA reconstructs the predicted main coefficients to the spectral dimension.
\begin{figure}[]
  \centering
  \includegraphics[width = 0.9\linewidth]{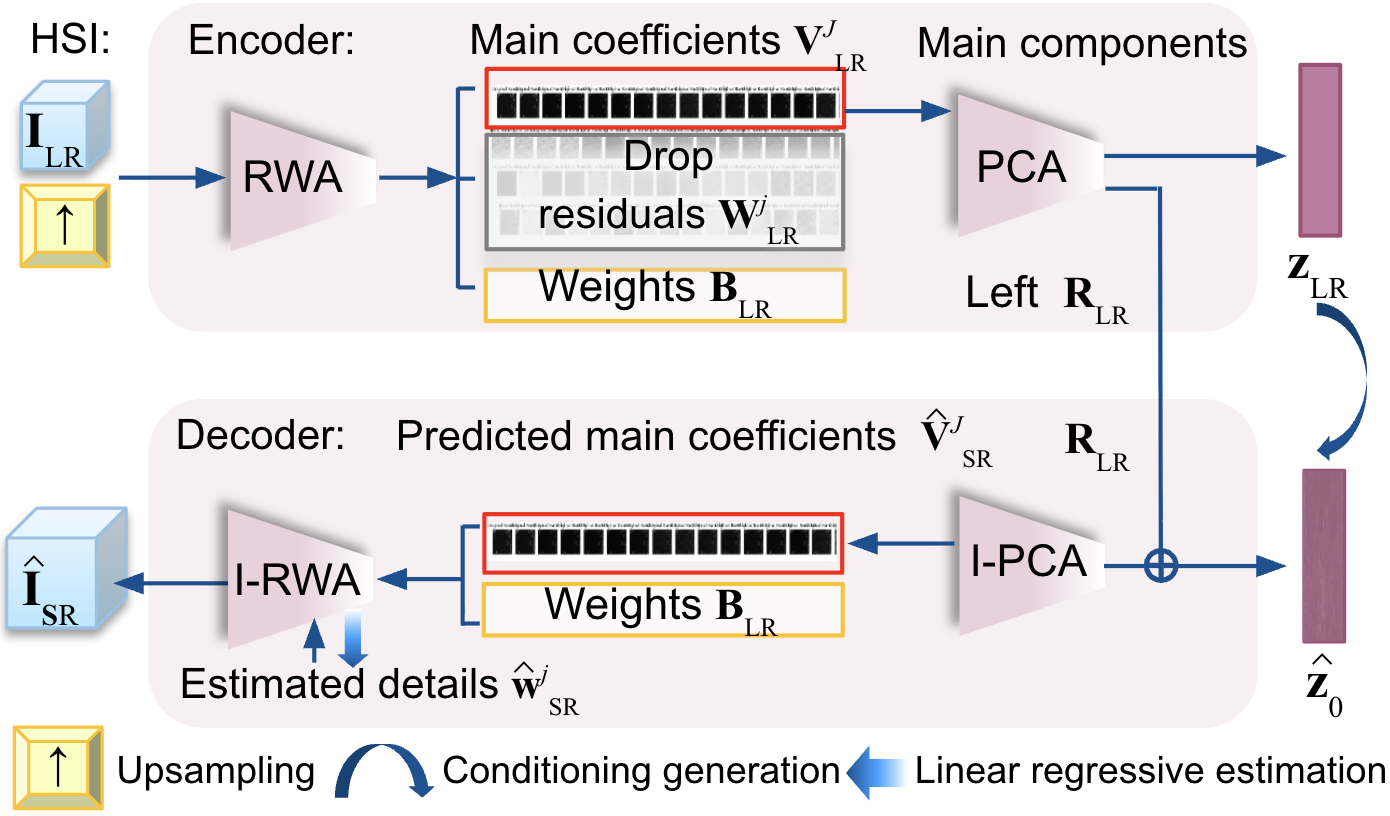}
  \caption{\new{Illustration of the wavelet-based encoder-decoder.}}
  \label{fig:ende}
\end{figure}

\subsection{Geometric enhanced diffusion process}
\label{sec:diff}
Hyperspectral image generation usually requires larger reverse sampling time steps. To solve this problem, we used EDM (Elucidating Diffusion Models) \cite{karras2022elucidatingedm} as our baseline model. EDM adds noise in one step in the training process. A probability flow ordinary differential equation (ODE) then continuously increases the noise level of the image when moving forward in time \cite{karras2022elucidatingedm}. Instead of using a discrete time step to add noise, we used a concrete number $\sigma$ to represent the strength of the noise: 
\begin{equation}
    \sigma \sim \textrm{exp}(\textrm{N}(P_{\textrm{mean}}=-1.2,P_{\textrm{std}}=1.2)),
\end{equation}
where $P_{\textrm{mean}}$ represents the mean value, $P_{\textrm{std}}$ is the standard deviation, and N is a Gaussian distribution.
 
The ``time variable" $t$ used in our model is a continuous variable, which has the advantage of mapping the noise scale to an approximately linear interval. In this way, the noise strength that will be added at the t moment can correspond to the size of t, and the relationship can be represented as:
\begin{equation}
    t=-\textrm{log}(\sigma_{t}).
\end{equation}
\paragraph{Edge-aware noise scheduler.} General diffusion models have an equal generation ability for each pixel. In the remote sensing scenario, we wanted to clarify the contour of buildings and other ground objects. Inspired by \cite{vandersanden2024edge}, we designed an edge-aware noise scheduler in our training stage to increase the generation ability of the diffusion model for the pixels around the edges. The edge is preserved during the forward diffusion process. The noise around the edge is smaller than the general noise:
\begin{equation}
    \textbf{z}_{\textrm{t}}=\textbf{z}_{0}+\sigma_{t}\epsilon\odot (1-\textbf{E} (1-\sigma _{\textrm{norm}}^{2}) \eta ),
\end{equation}
where $\textbf{z}_{\textrm{t}}$ represents the noisy features at moment t, $t\in \left ( 0,T \right )$, $\textbf{z}_{0}$ is the features from the ground truth, $\sigma_{t}$ represents the noise strength at moment t, $\epsilon\sim \textrm{N}(0 , 1)$ is random noise, \textbf{E} is the binary edge map obtained from the input image, $\sigma _{\textrm{norm}}$ is the normalized sigma at moment t, $\odot$ highlights that the operation is a matrix multiplication, and $\eta=0.5$ adjusts the perturbation strength influenced by the edge. 
\begin{figure}
  \centering
  \includegraphics[width = 1\linewidth]{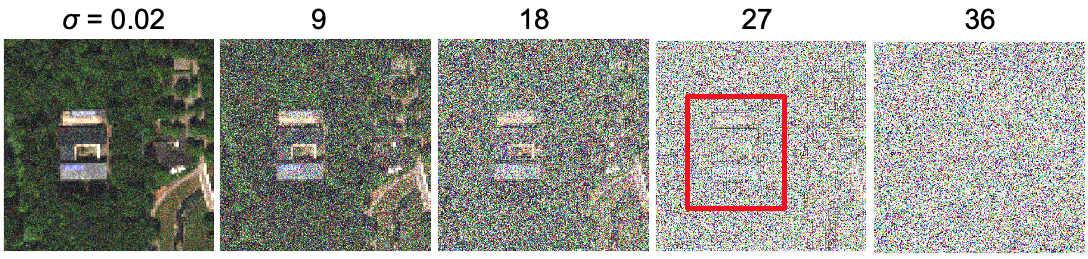}
  \caption{Edge perturbed noisy image over time.}
  \label{model}
\end{figure}

\paragraph{Mask controllable training and sampling.}
In our preliminary research, generating geometric objects without distortion was a big challenge. To address this, a mask is introduced as a condition that improves the ability to generate buildings. Segmentation is calculated from low-resolution RGB channels from hyperspectral images with a segment-anything model \cite{sam}. We used one minus the average of NDVI \cite{ndvi} as the value of the mask to highlight the attention of buildings. Let $S_s$ be the pixel number of the $s$ th segmentation region, then the value $M_s$ of the mask is defined as:
\begin{equation}
    M_s = 1 - \frac{1}{|S_s|} \sum_{(x, y) \in S_s} \mathrm{NDVI}_{\mathrm{norm}}(x, y), \mathrm{NDVI}_{\mathrm{norm}}\in [0,1].
\end{equation}

In the training stage, the predicted result $\mathbf{\hat{z}}_{0}$ is calculated in one step:
\begin{equation}
    \mathbf{\hat{z}}_{0} = f_{\theta }\left (  \mathbf{z}_{t}, \textbf{C}, \sigma _{t}\right ), \textbf{C}=[\mathbf{z}_{\textrm{LR}},\textbf{ M}],
\end{equation}
where $\mathbf{\hat{z}}_{0}$ represents the predicted features when t = 0; $f_{\theta }$ represents the objective function 3D U-Net with spectral fidelity enhancer (SFE) \cite{dong2021generativesfe}, as shown in figure \ref{fig:pipe}; $\mathbf{z}_{\textrm{LR}}$ represents the low-resolution condition; $\textbf{M}\in \left ( 0,1 \right )^{H\times W}$ is the mask condition; and \textbf{C} is the concatenation of all conditions.

During the sampling stage, DPM-Solver++ \cite{lu2022dpm} accelerates the generation by employing a second-order approximation to solve the underlying ODE, while utilizing adaptive time stepping to significantly reduce the number of function evaluations. In our sampler, $t\in [0,T]$ will be separated into $N$ steps. The step size is $\Delta t = t_{n+1}-t_{n}, n= 0,1, ...,N-1 $. The initial noisy image and noise strength at step n can then be calculated with:

\begin{equation}
    \mathbf{z}_{T}=\sigma _{T}\cdot \epsilon, 
\end{equation}
\begin{equation}
    \sigma _{n}=\left ( \sigma _{\textrm{max}}^{1/\rho } + \frac{n}{N-1}\left ( \sigma _{\textrm{min}}^{1/\rho }-\sigma _{\textrm{max}}^{1/\rho } \right )\right )^{\rho }, 
\end{equation}
where $\rho$ is the scheduling curvature parameter, and $\sigma _{\textrm{max/min}}$is the maximum/minimum noise strength. The n+1 step denoised features $\mathbf{z}_{n+1}$ can be calculated with:
\begin{equation}
    \gamma = -\frac{1}{2} \cdot \frac{t_{n+1} - t_n}{t_n - t_{n-1}},
\end{equation}
\begin{equation}
    \tilde{f}_\theta = (1 -\gamma) f_\theta(\hat{\mathbf{z}}_{n}, \mathbf{C},\sigma_{n}) +\gamma f_\theta(\hat{\mathbf{z}}_{n-1}, \mathbf{C},\sigma_{n-1}), 
\end{equation}
\begin{equation}
    \mathbf{z}_{n+1}=\frac{\sigma_{n+1}}{\sigma_{n}}\hat{\mathbf{z}}_{n}-\sigma_{n+1}(e^{-\Delta t}-1)\cdot \tilde{f}_\theta.
\end{equation}

\setlength{\tabcolsep}{1mm}
\begin{table*}[h!]
\centering
\fontsize{9pt}{9pt}\selectfont
\renewcommand{\arraystretch}{1} 
\begin{tabular}{ccccccccc}
\hline\hline
&\textbf{Metric} & \textbf{MCNet} & \textbf{MSDFormer} & \textbf{ESSAFormer} &  \textbf{DMGASR} &\textbf{HIR Diff} & \textbf{SNLSR} & \textbf{Ours} \\ \hline

\multirow{7}{*}{(a)} 
& PSNR↑  & 28.300$\pm$0.0480 & 28.284$\pm$0.0000 & 27.483$\pm$0.1392 &  26.986$\pm$0.2052 &24.833$\pm$0.1079 & 28.531$\pm$0.0001 & \textbf{28.863$\pm$0.2940} \\
& SSIM↑  & 0.6658$\pm$0.0025 & 0.6592$\pm$0.0000 & 0.5915$\pm$0.0191 &  0.5831$\pm$0.0118 &0.6401$\pm$0.0024 & 0.6718$\pm$0.0000 & \textbf{0.7104$\pm$0.0212} \\
& SAM↓   & 8.3332$\pm$0.1243 & 8.7442$\pm$0.0000 & 9.2114$\pm$0.2482 &  11.340$\pm$0.0743 &8.9538$\pm$0.0053 & \textbf{7.8911$\pm$0.0000} & 8.4283$\pm$0.3073 \\
& CC↑    & 0.7440$\pm$0.0000 & 0.7645$\pm$0.0000 & 0.7374$\pm$0.0004 &  0.6767$\pm$0.0128 &0.7543$\pm$0.0011 & 0.7527$\pm$0.0000 & \textbf{0.7945$\pm$0.0165} \\
& RMSE↓  & 0.0557$\pm$0.0002 & 0.0544$\pm$0.0000 & 0.0560$\pm$0.0002 &  0.0627$\pm$0.0006 &0.0810$\pm$0.0015 & 0.0552$\pm$0.0000 & \textbf{0.0548$\pm$0.0030} \\
& FID↓   & 116.14$\pm$0.0856 & 103.74$\pm$0.0000 & 97.438$\pm$14.779 &  49.026$\pm$2.3984 &50.596$\pm$1.0436 & 125.75$\pm$0.0691 & \textbf{44.464$\pm$17.627} \\
& LV↑    & 0.0004$\pm$0.0000 & 0.0004$\pm$0.0000 & 0.0004$\pm$0.0000 &  0.0037$\pm$0.0004 &0.0021$\pm$0.0002 & 0.0003$\pm$0.0000 & \textbf{0.0041$\pm$0.0022} \\ 

\multirow{7}{*}{(b)} 
& PSNR↑  & 24.216$\pm$0.0157 & 24.359$\pm$0.0000 & 24.103$\pm$0.0132 &  23.021$\pm$0.0146 &21.567$\pm$0.5351 & 24.305$\pm$0.0000 & \textbf{24.933$\pm$0.0079} \\
& SSIM↑  & 0.5355$\pm$0.0008 & 0.5536$\pm$0.0000 & 0.5210$\pm$0.0010 &  0.4925$\pm$0.0025 &0.4987$\pm$0.0133 & 0.5404$\pm$0.0000 & \textbf{0.6337$\pm$0.0106} \\
& SAM↓   & 11.663$\pm$0.0482 & 11.912$\pm$0.0000 & 12.166$\pm$0.0156 &  16.158$\pm$0.0117 &12.348$\pm$0.1154 & 11.418$\pm$0.0001 & \textbf{11.323$\pm$0.0456} \\
& CC↑    & 0.7050$\pm$0.0004 & 0.7238$\pm$0.0000 & 0.7077$\pm$0.0004 &  0.6575$\pm$0.0007 &0.7347$\pm$0.0003 & 0.7102$\pm$0.0000 & \textbf{0.7771$\pm$0.0003} \\
& RMSE↓  & 0.0685$\pm$0.0001 & 0.0669$\pm$0.0000 & 0.0682$\pm$0.0002 &  0.0779$\pm$0.0011 &0.0940$\pm$0.0061 & 0.0680$\pm$0.0000 & \textbf{0.0668$\pm$0.0020} \\
& FID↓   & 257.45$\pm$0.1949 & 272.78$\pm$0.0000 & 288.85$\pm$7.1990 &  120.06$\pm$11.769 &375.96$\pm$23.877 & 267.88$\pm$0.0032 & \textbf{64.333$\pm$4.2810} \\
& LV↑    & 0.0007$\pm$0.0000 & 0.0006$\pm$0.0000 & 0.0007$\pm$0.0000 &  0.0034$\pm$0.0005 &0.0005$\pm$0.0000 & 0.0005$\pm$0.0000 & \textbf{0.0087$\pm$0.0003} \\ 

\multirow{7}{*}{(c)} 
& PSNR↑  & 33.389$\pm$0.2641 & 28.709$\pm$0.0000 & 25.504$\pm$0.1340 &  32.864$\pm$0.2049 &34.473$\pm$0.0069 & 35.734$\pm$0.0000 & \textbf{35.837$\pm$0.1176} \\
& SSIM↑  & 0.7441$\pm$0.0029 & 0.4766$\pm$0.0000 & 0.4120$\pm$0.0312 &  0.6802$\pm$0.0159 &0.7362$\pm$0.0017 & 0.7525$\pm$0.0000 & \textbf{0.7747$\pm$0.0045} \\
& SAM↓   & 8.5500$\pm$0.0896 & 12.213$\pm$0.0000 & 18.724$\pm$0.5899 &  11.476$\pm$0.3843 &8.3601$\pm$0.0446 & 7.6613$\pm$0.0000 & \textbf{7.4735$\pm$0.0532} \\
& CC↑    & 0.6495$\pm$0.0145 & 0.6300$\pm$0.0000 & 0.6326$\pm$0.0090 &  0.5001$\pm$0.0161 &0.7102$\pm$0.0001 & 0.7733$\pm$0.0000 & \textbf{0.7906$\pm$0.0055} \\
& RMSE↓  & 0.0476$\pm$0.0006 & 0.0525$\pm$0.0000 & 0.0690$\pm$0.0006 &  0.0542$\pm$0.0013 &\textbf{0.0420$\pm$0.0001} & 0.0471$\pm$0.0000 & 0.0468$\pm$0.0006 \\
& FID↓   & 464.13$\pm$5.9021 & 738.62$\pm$0.0000 & 701.35$\pm$16.290 &  245.38$\pm$63.176 &363.23$\pm$6.9705 & 470.34$\pm$0.0000 & \textbf{238.12$\pm$16.970} \\
& LV↑    & 0.0003$\pm$0.0000 & 0.0003$\pm$0.0000 & 0.0010$\pm$0.0000 &  \underline{0.0031$\pm$0.0015} &0.0002$\pm$0.0000 & 0.0002$\pm$0.0000 & \textbf{0.0011$\pm$0.0000} \\ \hline
\multirow{4}{*}{(d)} 
&Tr time (s) &\num{1.33e4} & \num{3.99e4} & \num{7.65e4} &  \num{3.16e5} &-- & \num{2.80e4}  & \num{3.10e5} \\
&Te time (s) & 18.13 & 10.40 & 7.98 &  334.00 &212.90 & 4.10 & 28.70 \\
&NFE & $ 256^{2}$ & $ 256^{2}$ & $ 256^{2}$ &  $20\times8$ &20 & $ 256^{2}$ & 50 \\
&Model size & 6.50 MB & 57.7 MB & 3.70 MB &  1.18 GB &1.56 GB & 7.70 MB & 4.55 GB \\ \hline \hline
\end{tabular}
\caption{Quantitative comparison with SOTA SR models of PSNR, SSIM, SAM, CC, RMSE, FID, and LV on (a) MDAS sample 1, (b) MDAS sample 2, and (c) WDC dataset. (d) Model efficiency was evaluated with the training/testing time, number of function evaluations (NFE), and model size.
(Best performance value is highlighted in bold. Noise-affected values are underlined.)}
\label{tab:table1}
\end{table*}

\begin{figure*}[!ht]
  \centering
  \includegraphics[width = 0.97\linewidth]{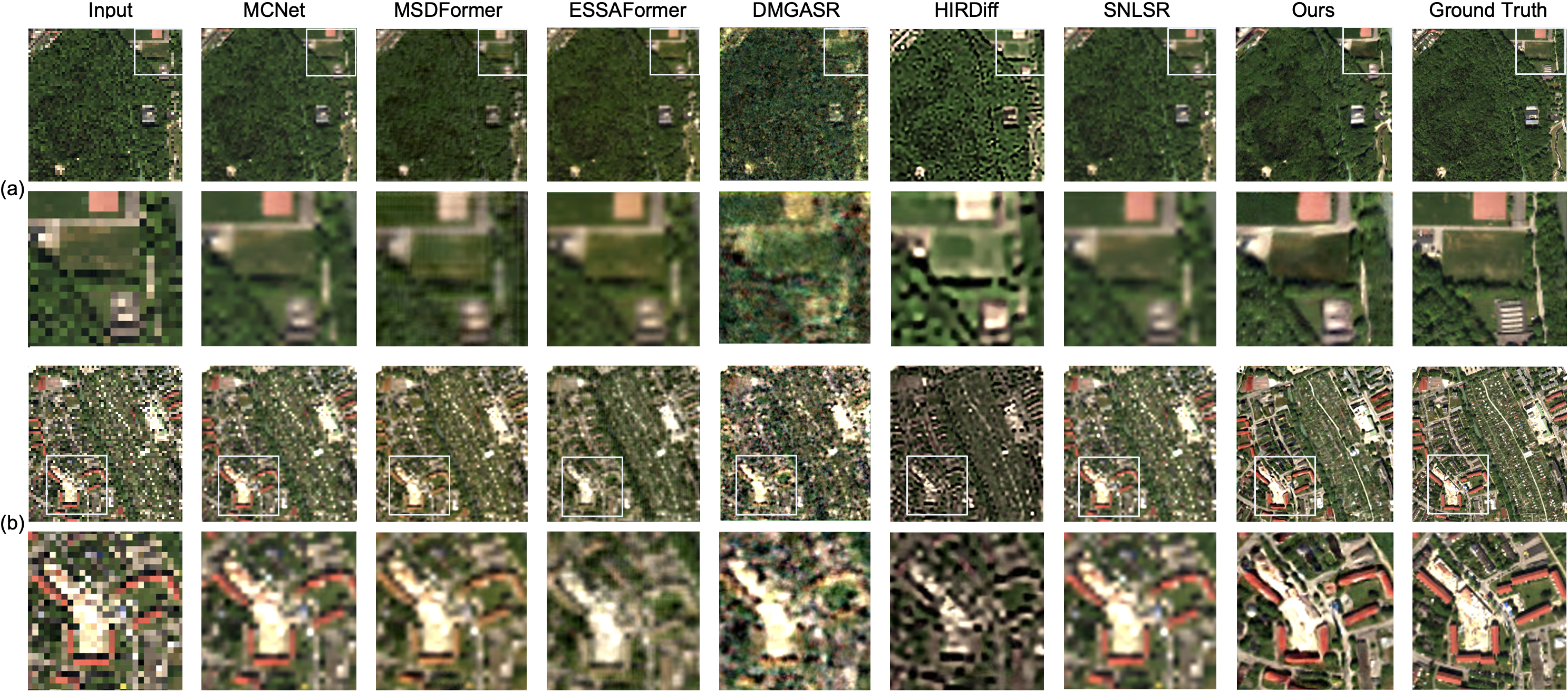}
  \includegraphics[width = 0.97\linewidth]{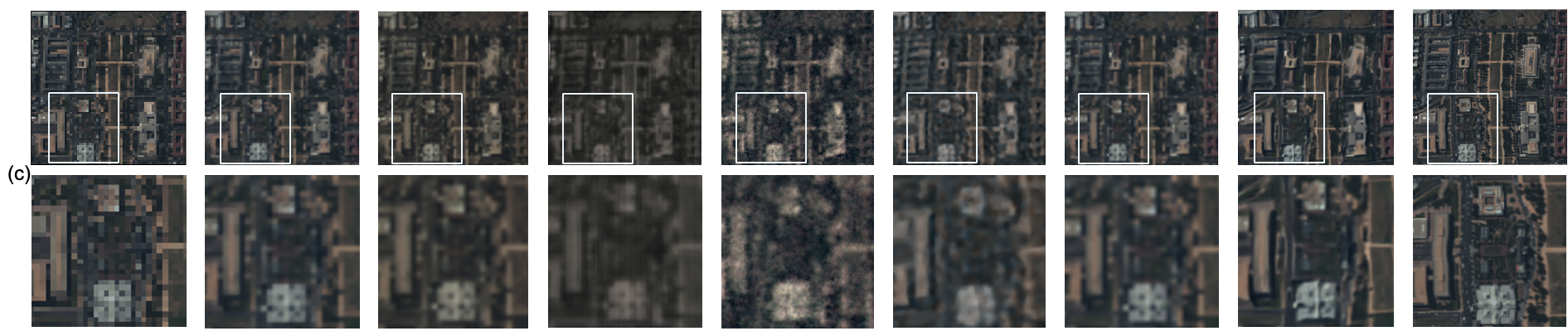}
  \caption{4-times visual comparisons with SOTA SR models on (a) MDAS sample 1, (b) MDAS sample 2, and (c) WDC dataset.}
  \label{fig:result}
\end{figure*}
\begin{figure*}[!ht]
  \centering
  \includegraphics[width = 1.0\linewidth]{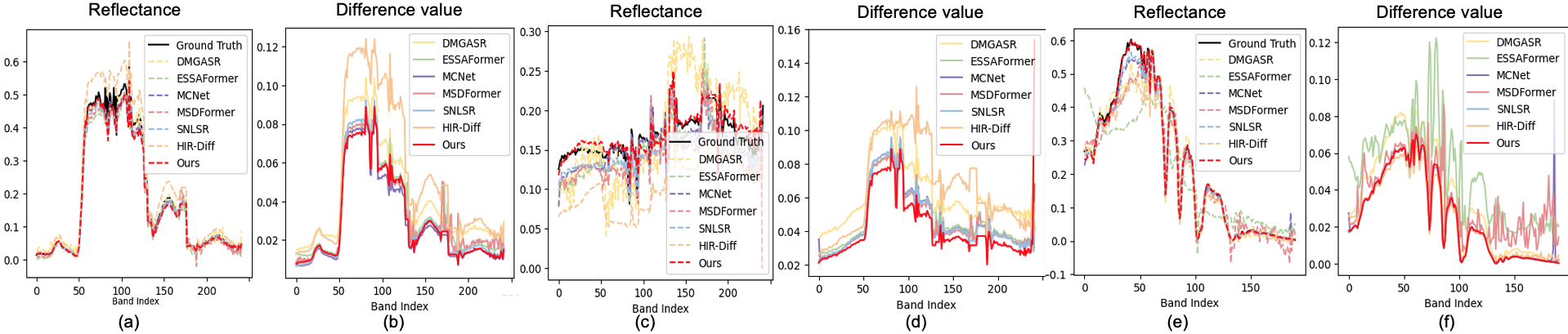}
  \caption{Spectral profile of a random pixel and mean difference value in each band of (a–b) MDAS sample 1, (c–d) MDAS sample 2, and (e–f) WDC dataset.}
  \label{spectral}
\end{figure*}
\begin{figure*}[!ht]
  \centering
  \includegraphics[width = 1\linewidth]{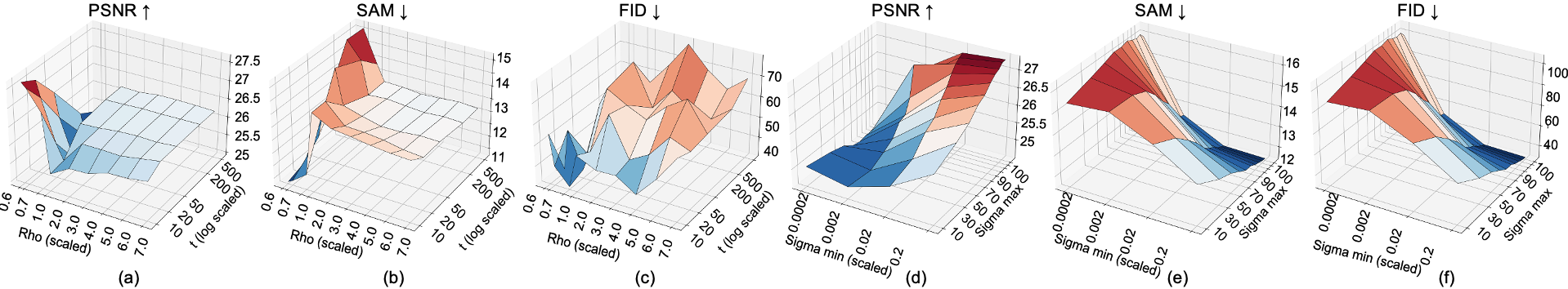}
  \caption{Qualitative comparison with different numbers of $\rho$ and time steps for (a) PSNR, (b) SAM, and (c) FID. Qualitative comparison with different numbers of $\sigma_{\textrm{max}}$ and $\sigma_{\textrm{min}}$  for (d) PSNR, (e) SAM, and (f) FID on validation dataset.}
  \label{fig:hyperpa}
\end{figure*}
\begin{figure}[]
  \centering
  \includegraphics[width = 1\linewidth]{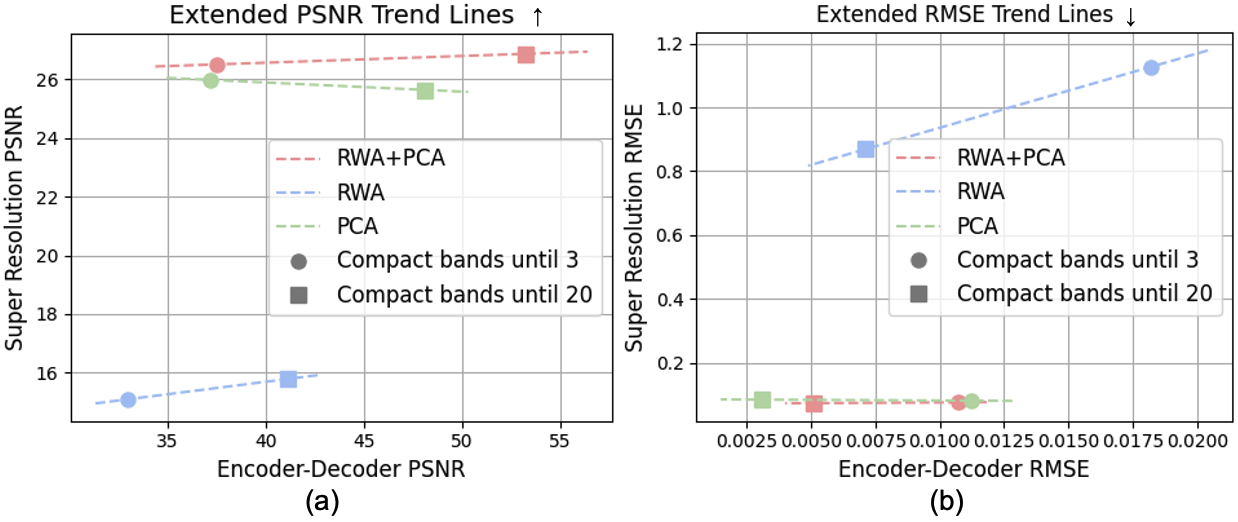}
  \caption{Encoder–decoder reconstruction quality compared with the performance of the whole SR process for the validation dataset.}
  \label{fig:endere}
\end{figure}
\subsection{Multi-level loss function}
Multi-level loss equations, such as equation \ref{eq:loss}, can ensure that the generated image is accurate in all aspects. 
\begin{equation}
    \label{eq:loss}
     \mathcal{L}= \lambda (t)\cdot (\lambda _{1}\mathcal{L} _{\textrm{pixel}}+\lambda _{2}\mathcal{L} _{\textrm{perc}}+\lambda _{3}\mathcal{L} _{\textrm{grad}}),
\end{equation}
To balance the convergence speed, we set $\lambda_{1} =0.8,\lambda_{2} =0.1,\lambda_{3} =0.1$. \new{$\lambda (t)$ indicates the loss weighting based on t.} Considering pixel loss can ensure that the absolute value of the spectral information of each pixel is accurate. Here we use the combination of L2 norm loss and Spectral Angle Mapper (SAM) \cite{yuhas1992sam} loss:
\begin{equation}
    \mathcal{L} _{\textrm{pixel}} = (\left\| \textbf{z}_{0}-\hat{\textbf{z}}_{0}\right\|^{2} + \textrm{SAM}(\textbf{z}_{0},\hat{\textbf{z}}_{0}))/2.
\end{equation}
Perceptron loss \cite{johnson2016perceptual} can ensure that the generated image is similar in high-level feature space: 
\begin{equation}
    \mathcal{L} _{\textrm{perc}} = \left\| \phi \textrm{VGG}(\hat{\textbf{z}}_{0})-\phi \textrm{VGG}(\textbf{z}_{0})\right\|_{2}^{2}.
\end{equation}
Gradient loss \cite{lu2022single} can ensure that the image gradient information is consistent with the real image. The images generated by DPM Solver++ have high-contrast characteristics. Gradient loss is defined as follows:

\begin{equation}
    \mathcal{L} \textrm{grad} = \frac{1}{2} (\left\| \bigtriangledown _{x}\hat{\textbf{z}}_{0}-\bigtriangledown _{x}\textbf{z}_{0}\right\|^{1}+\left\| \bigtriangledown _{y}\hat{\textbf{z}}_{0}-\bigtriangledown _{y}\textbf{z}_{0}\right\|^{1}),
\end{equation}
where $\bigtriangledown_{x/y}$ represents the gradient of the image in the x/y direction.
\section{Experiments}
\label{sec:exper}
\subsection{Datasets and implementation details}
\label{sec:dataset}
A group of EeteS simulated EnMap hyperspectral data, called the EnMap Campaign (realistic EnMAP-like high-resolution data), was used for training. EnMap is a hyperspectral satellite managed by the DLR Earth Observation Center, offering 30 m spatial resolution since June 2022. Additionally, the EnMAP Campaign Portal (access via supplementary material) captures aerial hyperspectral imagery and simulates EnMap-like data, boasting a spatial resolution of 2.5-4 m and containing data from 2009 to 2016. We gathered 8000 pairs of 256 × 256 × 242 patches of super-resolution HSI images and aligned them with 4-times downsampled images from EnMap Campaign and MDAS \cite{mdashu2023}. This dataset covers 15 cities in Europe and the Americas, representing diverse ground objects. Among 8,000 pairs, one pair was split into the validation dataset and used for parameter selection, and a group of ablation study experiments. Two pairs are selected for EnMap simulation testing. The WDC \cite{wdc} dataset is used as one of the test datasets to see the transmission on other datasets.
We deployed the model on four NVIDIA A100 GPUs with a learning rate of \num{1e-4} and trained it for 200 epochs.
\subsection{\new{Quantitative} metrics}
\label{sec:metrics}
\textbf{Fidelity} was evaluated using two standard metrics: the Peak Signal-to-Noise Ratio (PSNR) \cite{huynh-psnr}, which measures the pixel-wise quality, and the Structural Similarity Index (SSIM) \cite{ssim}, which captures structural consistency. Both were averaged over spectral bands.
\textbf{Realism and clarity} were addressed using the Fréchet Inception Distance (FID) \cite{heusel2017fid}, noting that it is computed in an RGB-trained feature space and thus was only used for relative comparisons. We also included Local Variation (LV) \cite{pertuz2013lv} to evaluate the local texture sharpness. \textbf{Spectral accuracy} was measured via Spectral Angle Mapper (SAM) \cite{yuhas1992sam}, Cross-Correlation (CC), and Root Mean Square Error (RMSE). These metrics assess the angular, correlational, and pixel-wise spectral alignment.

\begin{table*}[t]
\centering
\fontsize{9pt}{9pt}\selectfont
\setlength{\tabcolsep}{1mm}
\begin{tabular}{ccccccccccccc}
\hline \hline
\textbf{Method} & \textbf{Baseline} & \textbf{A} & \textbf{B} & \textbf{C} & \textbf{D} & \textbf{E} & \textbf{F} & \textbf{G} & \textbf{H} & \textbf{I} & \textbf{J} & \textbf{Ours} \\ \hline
w/RWA & \multicolumn{1}{l}{} & \cellcolor[HTML]{EFEFEF}\checkmark & \multicolumn{1}{l}{\cellcolor[HTML]{EFEFEF}} & \checkmark & \checkmark & \checkmark & \checkmark & \checkmark & \checkmark & \checkmark & \checkmark & \checkmark \\
w/PCA & \multicolumn{1}{l}{} & \multicolumn{1}{l}{\cellcolor[HTML]{EFEFEF}} & \cellcolor[HTML]{EFEFEF}\checkmark & \checkmark & \checkmark & \checkmark & \checkmark & \checkmark & \checkmark & \checkmark & \checkmark& \checkmark \\
w/Mask & \multicolumn{1}{l}{} & \checkmark & \checkmark & \cellcolor[HTML]{EFEFEF}\checkmark & \multicolumn{1}{l}{\cellcolor[HTML]{EFEFEF}} & \checkmark & \checkmark & \checkmark & \checkmark & \checkmark & \checkmark& \checkmark \\
w/Edge & \multicolumn{1}{l}{} & \checkmark & \checkmark & \multicolumn{1}{l}{\cellcolor[HTML]{EFEFEF}} & \cellcolor[HTML]{EFEFEF}\checkmark & \checkmark & \checkmark & \checkmark & \checkmark & \checkmark & \cellcolor[HTML]{EFEFEF} Inverse & \checkmark \\
w/L pix & \checkmark & \checkmark & \checkmark & \checkmark & \checkmark & \cellcolor[HTML]{EFEFEF}\checkmark & \multicolumn{1}{l}{\cellcolor[HTML]{EFEFEF}} & \multicolumn{1}{l}{\cellcolor[HTML]{EFEFEF}} & \checkmark & \checkmark & \checkmark& \checkmark \\
w/L perc & \multicolumn{1}{l}{} & \checkmark & \checkmark & \checkmark & \checkmark & \multicolumn{1}{l}{\cellcolor[HTML]{EFEFEF}} & \cellcolor[HTML]{EFEFEF}\checkmark & \multicolumn{1}{l}{\cellcolor[HTML]{EFEFEF}} & \checkmark & \checkmark & \checkmark& \checkmark \\
w/L geo & \multicolumn{1}{l}{} & \checkmark & \checkmark & \checkmark & \checkmark & \multicolumn{1}{l}{\cellcolor[HTML]{EFEFEF}} & \multicolumn{1}{l}{\cellcolor[HTML]{EFEFEF}} & \cellcolor[HTML]{EFEFEF}\checkmark & \checkmark & \checkmark & \checkmark& \checkmark \\
w/Unet3D & \multicolumn{1}{l}{} & \checkmark & \checkmark & \checkmark & \checkmark & \checkmark & \checkmark & \checkmark & \cellcolor[HTML]{EFEFEF}\checkmark & \multicolumn{1}{l}{\cellcolor[HTML]{EFEFEF}} & \checkmark& \checkmark \\
w/SFE & \multicolumn{1}{l}{} & \checkmark & \checkmark & \checkmark & \checkmark & \checkmark & \checkmark & \checkmark & \multicolumn{1}{l}{\cellcolor[HTML]{EFEFEF}} & \cellcolor[HTML]{EFEFEF}\checkmark & \checkmark& \checkmark \\ 
PSNR↑ & 2.0476 & 15.788 & 25.640 & 26.579 & 26.681 & 26.101 & 21.664 & 26.467 & 26.388 & 26.181 &26.8713& \textbf{27.013} \\
SSIM↑ & -0.0150 & -0.0173 & 0.5267 & 0.6503 & 0.6530 & 0.6140 & 0.1580 & 0.6443 & 0.6240 & 0.6131 & 0.6667&\textbf{0.6573} \\
SAM ↓ & 124.15 & 85.239 & 15.125 & 11.766 & 12.160 & 12.804 & 25.528 & 12.445 & 12.788 & 14.085 & 11.7688&\textbf{11.501} \\
CC↑ & 0.2643 & 0.4763 & 0.5530 & 0.6803 & 0.6882 & 0.6441 & 0.0145 & 0.6741 & 0.6681 & 0.6500 &0.6999 &\textbf{0.7008} \\
RMSE ↓ & 1.1879 & 0.8687 & 0.0850 & 0.0758 & 0.0749 & 0.0804 & 0.1342 & 0.0769 & 0.0783 & 0.0803 & 0.0739&\textbf{0.0726} \\
FID ↓ & 5019.1 & 484.16 & 83.627 & 43.445 & 36.269 & 40.303 & 701.94 & 40.195 & 47.475 & 65.985 & 34.9402&\textbf{30.110} \\
LV↑ & 7.5344 & 0.6231 & 0.0160 & 0.0079 & 0.0077 & 0.0096 & 0.1684 & 0.0093 & 0.0074 & 0.0089 & 0.0080&\textbf{0.0083} \\ \hline \hline
\end{tabular}
\caption{Ablation study. Quantitative comparison for the effect of each module on the validation dataset. Results are averaged over 4 runs. The baseline used an EDM backbone and DPM-Solver++ sampler on 242 bands. A: no PCA in the encoder; B: no RWA in the encoder; C: no edge perturbation; D: no mask conditioning; E/F/G: retained pixel/perceptual/geometric loss; H: no spectral fidelity module in the encoder; I: used 2D U-Net instead of 3D; \new{J: more noise on edge \cite{zhang2025uncertaintyguidedperturbationimagesuperresolution}.}}
\label{tab:table2}
\end{table*}

\subsection{State-of-the-art image generation}
\label{sec:result}
We evaluated the performance and efficiency of our model by comparing it with the six SOTA models: MCNet \cite{mcnet}, MSDFormer \cite{chen2023msdformer}, ESSAFormer \cite{zhang2023essaformer}, DMGASR \cite{wang2024dmgasr}, HIR Diff \cite{pang2024hirdiff}, and SNLSR \cite{snlsr}. We trained these models on our dataset with the implementation details provided in each paper, except for the unsupervised method HIR Diff. The HIR Diff model provides a pre-trained checkpoint based on their fully prepared HSI dataset.
\begin{figure}[]
  \centering
  \includegraphics[width = 0.84\linewidth]{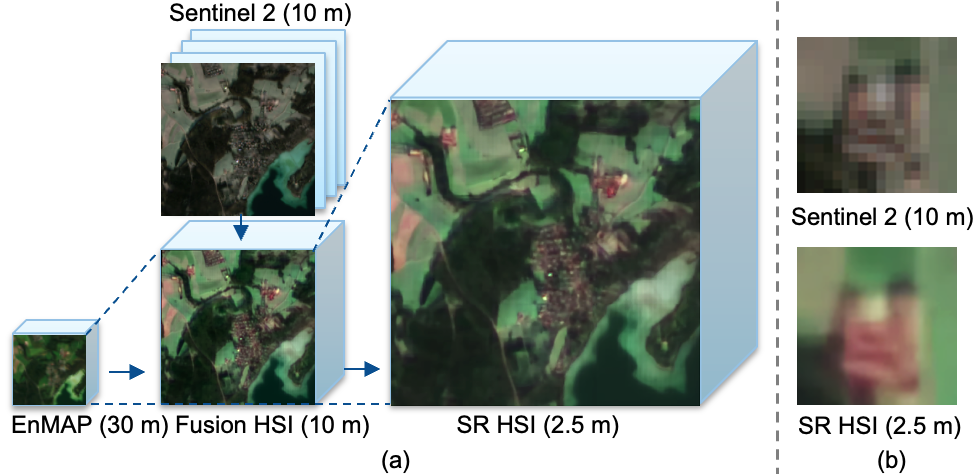}
  \caption{(a)Visualization of EnMap, Sentinel 2, Fusion image, Super resolution image. (b) Effect of super-resolution.}
  \label{fig:enmap}
\end{figure}
Our model demonstrated strong performance in generating medium to large-scale ground objects. In Figure \ref{fig:result} and Table \ref{tab:table1}, we can see that both the visualization result and quantitative result outperformed the other models. The WDC dataset result shows our model can adapt to another dataset with a different spectral profile. However, our model may struggle with accuracy when the input conditional image lacks sufficient semantic information; for example, the rooftop of the MDAS sample 1 reconstructed image. 
\paragraph{Effect of the encoder-decoder alone vs. with super-resolution.}
We compared our encoder-decoder (RWA + PCA) against RWA-only and PCA-only baselines (Figure \ref{fig:endere}), under 3- and 20-band compression. Without super-resolution, our encoder-decoder achieved almost lossless reconstruction with a PSNR up to 56. With super-resolution, using RWA + PCA compression to 20 bands outperformed all the baselines, demonstrating strong spectral compaction and generation. As shown, retaining more bands improves the quality but increases the cost. We chose 20 bands to ensure a balance between performance and efficiency.
\label{sec:para}
\paragraph{Effect of {$\rho$}, number of timesteps $N$, $\sigma_{\textrm{max}}$ and $\sigma_{\textrm{min}}$ on sampling.}
The EDM noise schedule is shaped by $\rho$, which controls how noise levels decay from $\sigma_{\text{max}}$ to $\sigma_{\text{min}}$ via $N$ steps. Higher $\rho$ sharpens early denoising but increases the randomness in later steps, potentially degrading spectral consistency. We found that lower $\rho$ values [0.6–0.7] yielded smoother noise schedules via 50 steps, which also better preserve spectral fidelity (Figure \ref{fig:hyperpa}). 
We also studied the effects of $\sigma_{\text{max}}$ and $\sigma_{\text{min}}$. A larger $\sigma_{\text{max}}$ allows more diversity but risks over-noising; a smaller one limits the detail. We set $\sigma_{\text{max}}=80$ for a balance. For $\sigma_{\text{min}}$, smaller values extend the denoising phase, improving the detail but increasing artifacts. We found the best results when $\sigma_{\text{min}} \in [0.02, 0.2]$.

\paragraph{Real-world application.} We tried to combine EnMAP and Sentinel-2 imagery to 10 m resolution hyperspectral data via the unsupervised method HySure \cite{hysure}. Leveraging our 4-times super-resolution generation model, GEWDiff, we could finally produce EnMAP hyperspectral images with a 2.5-meter resolution. The no-reference image quality assessment MetaIQA \cite{zhu2020metaiqadeepmetalearningnoreference} was improved from 0.1997 to 0.2029 (Figure \ref{fig:enmap} (b)). 

\subsection{Ablation study}
The results of our ablation studies, presented in Table \ref{tab:table2}, offer significant insights into the contributions of various components of the GEWDiff model. The use of a suitable encoder plays a crucial role in our model design. The multi-level loss function achieves better performance than an L2 loss. The design of a 3D objective function, U-Net, and its spectral fidelity enhancer makes progress toward stabilizing the results. The geometric enhancement strategies, such as edge perturbation and mask conditioning, did not show a significant improvement in the global metrics. However, some effects could be observed from Figure \ref{fig:result}, whereby the edges are clear, and there was no obvious building distortion. 
\section{Conclusion}
\label{headings}
We proposed GEWDiff, which improves the spatial resolution of hyperspectral images by a factor of 4. Our method integrates wavelet-domain transforms and geometric priors to effectively preserve both spectral fidelity and spatial textures while accelerating convergence. The experimental results showed that GEWDiff outperformed the current SOTA baselines. One limitation of GEWDiff is that the result relies too much on the input conditions. This limitation could be addressed in future work through the integration of classifier-free guidance, which would enable the model to better generalize under weak or ambiguous conditioning. \new{Moreover, we would also contribute to model distillation for further lightweight alternatives.}

\section{Acknowledgments}
This manuscript has been accepted for publication in AAAI 2026. Please refer to the original version via [link: \url{https://aaai.org/conference/aaai/aaai-26/}].

This work was supported in part by [the place to write support project]; and in part by Munich Center for Machine Learning. The authors sincerely thank GFZ Helmholtz-Zentrum for providing the EnMAP Campaign datasets \cite{enmap1,enmap2,enmap3,enmap4,enmap7,enmap8,enmap9,enmap10,enmap12,enmap13,enmap15} used in this paper, and Dr. Jingliang Hu for providing the MDAS dataset.

\bibliography{aaai2026}

\appendix

\section{Appendix}
\begin{figure}[!ht]
  \centering
  \includegraphics[width = 1.0\linewidth]{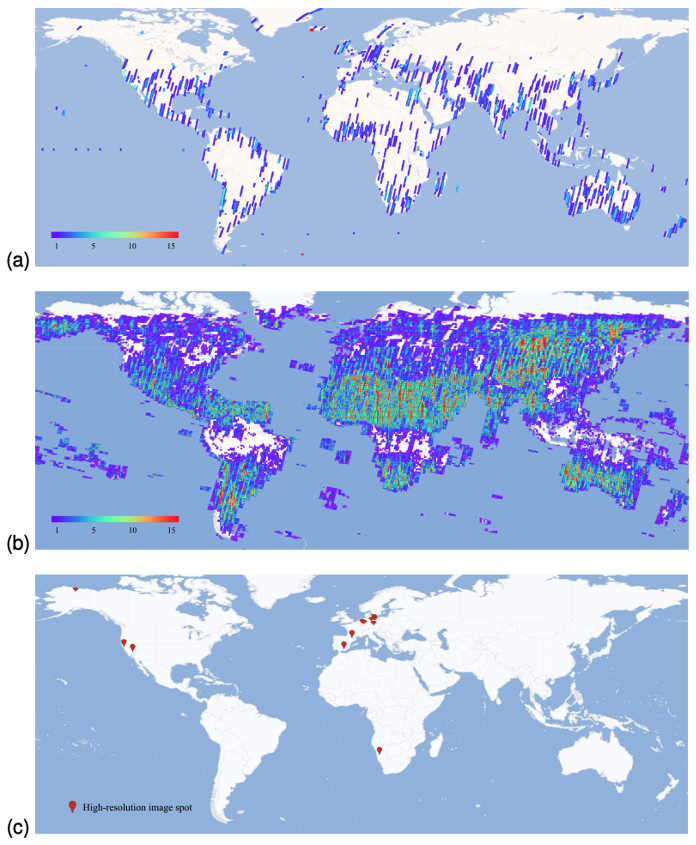}
  \caption{Satellite trajectory and number of repeat times for (a) EnMap ($GSD=30m$) and (b) Sentinel-2 ($GSD=10m$) satellite from 01.2024 to 03.2024. (Sentinel-2 images are filtered by cloud coverage $< 10\%$) (c)Location of high-resolution EnMap Campaign ($GSD\approx2.5m$) hyperspectral airborne images.}
  \label{fig:cover}
\end{figure}

There is a growing demand for high-quality, global-scale hyperspectral data. As shown in Figure \ref{fig:cover}, we compare the global coverage of EnMap data, Sentinel-2 data, and high-resolution airborne hyperspectral imagery. It is evident that higher data quality typically comes at the expense of reduced spatial coverage. In this work, we aim to develop a cost-efficient AI-based model that enhances the quality of hyperspectral satellite imagery (e.g., EnMap), thereby making high-quality data more accessible for downstream applications such as environmental monitoring, land cover classification, and precision agriculture.
\subsection{Research review for HSI super-resolution.} A Mixed Convolutional Network (MCNet) \cite{mcnet} is proposed for hyperspectral image super-resolution (SR), which introduces a Mixed Convolutional Module (MCM) with 3D CNNs. A novel deep learning method named PDE-Net is proposed \cite{hou2022deep} to formulate hyperspectral (HS) image embedding as an approximation of the posterior distribution of spatial-spectral features. 
Generative adversarial networks (GANs) have gained attention as a promising class of generative models for hyperspectral image super-resolution. A Latent-Encoder GAN (LE-GAN) \cite{legan} introduces a latent encoder to map spectral-spatial features into a latent space. A GAN-based method with band attention \cite{bagan} leverages full-band input and spectral-spatial constraints to reduce texture blur and spectral distortion. However, they often suffer from limitations such as mode collapse and low sample diversity, which can limit their ability to fully capture the complex distribution of hyperspectral data. 
Notable examples include ESSAformer \cite{zhang2023essaformer}, which introduces enhanced spectral-spatial attention mechanisms; MSDFormer \cite{chen2023msdformer} leverages multi-scale dynamic fusion; DSTrans \cite{yu2023dstrans}, which incorporates dynamic spectral transformers; and EigenSR \cite{su2025eigensr}, which explores low-rank representations within a transformer framework. Transformers show architectures' growing versatility and effectiveness in modeling complex hyperspectral data. While these models effectively maintain spectral consistency across bands, they often fail to generate fine-grained spatial details and high-frequency textures that are critical for accurate hyperspectral reconstruction.

\begin{table*}[!ht]
\centering
\fontsize{9pt}{9pt}\selectfont
\label{tab:data_summary}
\begin{tabular}{ccccc}
\hline \hline
Name & Application & Sensor & Product-Level & Date \\ \hline
Gerobstein (Germany) & Forest & HySpec & L2 & 2016-09-08 \\ 
Tonlik Lake (Alaska, USA) & Vegetation & AISA (Eagle) & L2 & 2016-08-27 \\ 
Drenmin (Germany) & Soil & HySpex VNIR-1600 & L2 & 2015-10-01 \\ 
 &  & HySpex SWIR320m-e &  &  \\ 
Namäla & Soil & HySpex VNIR-1600 & L2 & 2015-06-06 \\ 
 &  & HySpex SWIR320m-e &  &  \\ 
Donnersberg (DE) & Forest & HySpex VNIR-1600 & L2 & 2014-07-02 \\ 
 &  & HySpex SWIR320m-e &  &  \\ 
Mountain Pass, USA & Geology & AVIRIS-NG & L2 & 2014-06-21 \\ 
Redalquillar, Spain & Geology & HyMap & L2 & 2014-06-15 \\ 
Hamrdick-Hochwald (DE) & Forest & HySpex VNIR-1600 & L2 & 2014-05-04 \\ 
 &  & HySpexc SWIR320m-e &  &  \\ 
San Francisco Bay Area (USA) & Vegetation & AVIRIS-classic & L2 & 2013-11-22 \\ 
San Francisco Bay Area (USA) & Vegetation & AVIRIS-classic & L2 & 2013-06-07 \\ 
San Francisco Bay Area (USA) & Vegetation & AVIRIS-classic & L2 & 2013-04-10 \\ 
Neurling (DE) & Agriculture & AVIS-3 & L2 & 2012-09-07 \\ 
Neurling (DE) & Agriculture & HySpex VNIR-1600 & L2 & 2012-08-11 \\ 
 &  & HySpex SWIR320m-e &  &  \\ 
MDAS (DE) & Urban/Urban-Rural & HySpex & L2 & 2018-5-7 \\ 
Neurling (DE) & Agriculture & AVIS-3 & L2 & 2012-06-15 \\ 
Neurling (DE) & Agriculture & AVIS-3 & L2 & 2012-05-24 \\ 
Köthen (DE) & Agriculture & asiaDual & L2 & 2012-05-23 \\ 
Neurling (DE) & Agriculture & HySpex VNIR-1600 & L2 & 2012-05-07 \\ 
 &  & HySpex SWIR320m-e &  &  \\ 
Neurling (DE) & Agriculture & AVIS-3 & L2 & 2012-04-27 \\ 
Neurling (DE) & Agriculture & APEX & L2 & 2011-09-09 \\ 
Isabean (ES) & Environmental gradients & asiaEagle & L2 & 2011-08-08 \\ 
Köthen (DE) & Agriculture & asiaDual &  L2 & 2011-06-26 \\ 
Köthen (DE) & Agriculture & asiaDual &  L2 & 2011-05-09 \\ 
Isabean (ES) & Environmental gradients & asiaEagle &  L2 & 2011-04-01 \\ 
Döbertizer Heide (DE) & Vegetation & HyMap &  L2 & 2009-08-19 \\ 
Berlin (DE) & Urban/Urban-Rural & HyMap &  L2 & 2009-08-19 \\ 
Berlin (DE) & Urban/Urban-Rural & HyMap &  L2 & 2009-08-19 \\ 
Neurling (DE) & Agriculture & HyMap &  L2 & 2009-07-26 \\ 
Steinbeißen (DE) & Agriculture & HyMap &  L2 & 2009-07-26 \\ 
Döbertizer Heide (DE) & Vegetation & HyMap &  L2 & 2008-08-06 \\ \hline\hline
\end{tabular}
\caption{Data Collection Summary}
\end{table*}
\subsection{Challenges in hyperspectral image generation.} Despite recent progress, hyperspectral image (HSI) generation still faces several practical and technical challenges, we explained the details for challenges and the solution we give in our paper:
\begin{itemize}
    \item \textbf{Computational cost. } Diffusion models such as DDPM \cite{ho2020ddpm} are memory-intensive, requiring up to 1.3 GB of GPU memory per spectral channel, making them hard to scale to high-bandwidth data. In contrast, our proposed method with 3D U-Net reduces this requirement to a maximum of 1.0 GB per channel. Additionally, we can provide the code for a lightweight version of the model with a 2D U-Net, which operates in a 3-channel latent space. This configuration supports a batch size of 16 without significant performance degradation and is fully trainable on a single NVIDIA RTX 3090 GPU.

    \item \textbf{Sampling speed.} Generating HSIs typically requires thousands of timesteps (e.g., 5000 for DDPM), which significantly slows down inference. Our proposed sampling scheduler reduces to 50 timesteps, which can be tested on CPU devices.

    \item \textbf{Low signal-to-noise ratio.} Maintaining strong signal integrity across many channels is difficult, especially under noise-aware training objectives. Hyperspectral image generation is particularly challenging due to the inherently low signal-to-noise ratio. To address this, we employ an encoder-decoder architecture that transforms the complex high-dimensional image space into a more compact and structured latent space.

    \item \textbf{Slow convergence. }Diffusion models tend to converge very slowly when processing a large number of channels simultaneously. To alleviate this issue, our encoder-decoder architecture effectively reduces the number of channels to 20 in the latent space.

    \item \textbf{Geometric distortion.} Preserving geometric correctness across spectral bands is critical but challenging, especially when using spatially-aware denoising processes. Conventional natural image generation models often struggle to capture the unique geometric and structural patterns required for remote sensing image synthesis. To address this, we propose a mask-controllable, edge-aware architecture that enhances geometric fidelity and reduces spatial distortion in the generated hyperspectral images.

    \item \textbf{Real-world applicability.} For satellite-based HSI (e.g., EnMap), a two-stage resolution enhancement is necessary. First, spatial resolution is increased to Sentinel-2 level (10 m) using a fusion model trained on EeteS [2] simulated data. Then, a diffusion model further upsamples it to 2.5 m, enabling high-resolution, global-scale hyperspectral generation.
\end{itemize}

\section{Dataset}
Hyperspectral aerial images that have been used in the framework are called EnMAP Campaign \cite{enmap1,enmap2,enmap3,enmap4,enmap5,enmap6,enmap7,enmap8,enmap9,enmap10,enmap11,enmap12,enmap13,enmap15}. It is a preparatory science program to support method and application development in the prelaunch phase of the EnMAP satellite mission. This project aims to use aerial HSI data to simulate EnMap images. We also used another EnMap-like aerial dataset called the MDAS dataset \cite{mdashu2023}. All datasets are shown in Table. 1.

We use Level-2 reflectance products with digital number (DN) values ranging from 0 to 10,000, and a scale factor of 10,000. Each sample consists of a low-resolution input patch of size 64 × 64 × 242 and a corresponding high-resolution ground truth patch of 256 × 256 × 242, resulting in 8000 paired samples for training and evaluation. We use the MDAS dataset as the reference spectral profile, as it has been preprocessed using the EeteS \cite{EeteS} tool to simulate EnMap spectral characteristics. To ensure consistency across datasets from different sensors, we apply nearest-neighbor interpolation to align their spectral profiles with that of MDAS.

\section{Methodology}
In this section, we include two paragraphs that were not shown in our paper for a better understanding.
\subsection{Information loss of wavelet-based encoder-decoder.} Our encoder-decoder is designed to compact hyperspectral images into a latent space by operating solely on the spectral dimension. However, minimizing information loss during this spectral compression is critical for accurate HSI reconstruction. In this section, we evaluate the effectiveness of our approach in achieving an almost lossless transformation, demonstrating its ability to preserve essential spectral information. 

RWA significantly improves the redundancy processing capability of DWT by introducing a regression model. During the inverse transformation, the detail coefficients are predicted from the main coefficients through the regression model, and the original image is reconstructed using the saved weights. The results of the wavelet transform are usually sparse, even though they compact the HSI efficiently. In contrast, the results of PCA are usually dense, and the principal components are orthogonal. PCA is more suitable for processing global features in the diffusion model. As a result, we combine two transformers.

\begin{figure}[!ht]
  \centering
  \includegraphics[width = 0.6\linewidth]{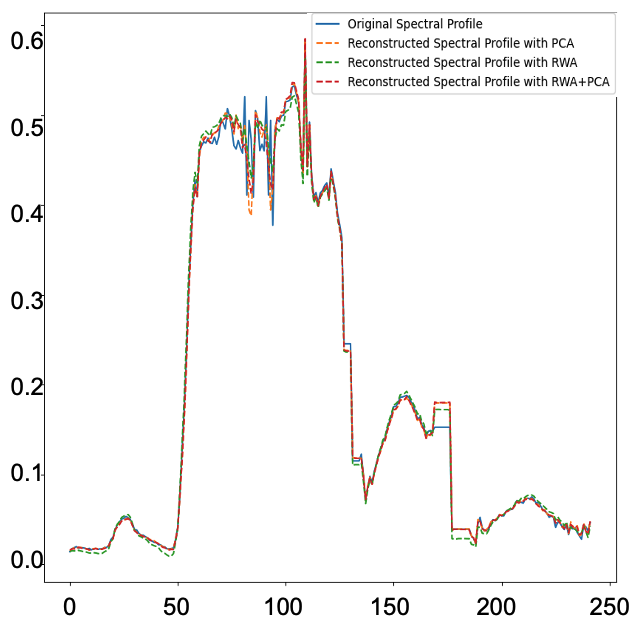}
  \caption{Spectral profile of pixel (125,125) recovered encoder  → decoder original vs. reconstructed by RWA, PCA and RWA+PCA.}
  \label{fig:endespe}
\end{figure}

In Figure \ref{fig:ende}, we present reconstructed images using RWA+PCA, RWA-only, and PCA-only methods in line (a). The compression to 3–4 bands is intentional to better visualize reconstruction errors. The RWA-only method struggles to recover full color information, while the PCA-only approach produces overly smooth, blurry results. In contrast, RWA+PCA yields clearer latent features (line (b)) and more faithful reconstructions. In Figure \ref{fig:endespe}, all methods demonstrate strong recovery of spectral profiles, even when compressed to just 3–4 bands. In practice, we compact to 20 bands to ensure an almost lossless spectral transformation.

Table. 2 presents a quantitative comparison of various RWA levels combined with different numbers of retained PCA components. The first row shows the final configuration used in our experiments, chosen to balance GPU memory usage and encoder-decoder efficiency.
\begin{figure}[!ht]
  \centering
  \includegraphics[width = 1.0\linewidth]{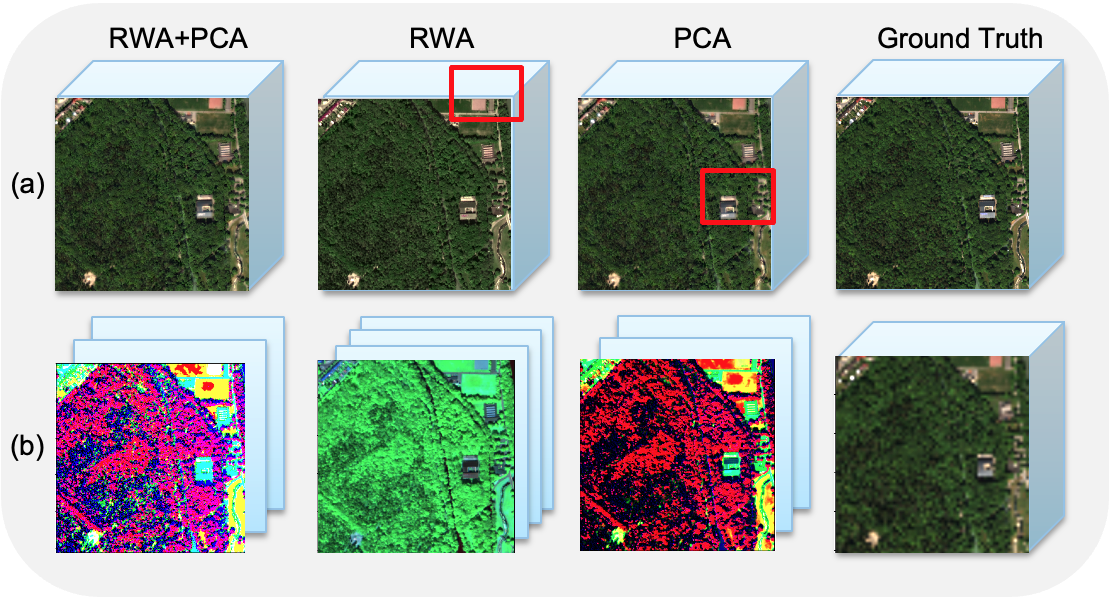}
  \caption{(a) Recovered encoder  → decoder  (RGB visualization), (b) Latent space (False color visualization for first three bands).}
  \label{fig:ende}
\end{figure}
\setlength{\tabcolsep}{2mm}
\begin{table*}[!ht]
\centering
\label{tab:factor}
\fontsize{9pt}{9pt}\selectfont
\begin{tabular}{cccccccc}
\hline\hline
\textbf{RWA compact until} & \textbf{PCA compact until} & \textbf{MPSNR↑} & \textbf{MSSIM↑} & \textbf{ERGAS↓} & \textbf{SAM↓} & \textbf{CrossCorrelation↑} & \textbf{RMSE↓} \\ \hline
121 bands \checkmark    & 20  bands\checkmark  & 53.26           & 0.9962          & 0.9505          & 1.7577        & 0.9982                     & 0.0051         \\
31 bands           & 20 bands           & 52.14           & 0.995           & 1.1208          & 2.3755        & 0.9974                     & 0.0068         \\
61 bands           & 20 bands          & 52.77           & 0.9956          & 0.9889          & 2.1529        & 0.9979                     & 0.0062         \\
61 bands           & 10 bands          & 49.81           & 0.9939          & 1.3837          & 2.3702        & 0.9967                     & 0.0068         \\
16 bands          & 10 bands          & 48.76           & 0.9933          & 1.4816          & 2.5506        & 0.9961                     & 0.0073         \\
121 bands         & 10 bands          & 49.42           & 0.9942          & 1.5159          & 2.0848        & 0.9965                     & 0.006          \\
31  bands         & 10 bands          & 50.01           & 0.9938          & 1.3397          & 2.4749        & 0.9967                     & 0.0071         \\
242 bands         & 10 bands          & 49.14           & 0.994           & 1.6179          & 1.8881        & 0.9962                     & 0.0055         \\
16  bands         & 6 bands           & 46.47           & 0.9898          & 1.8661          & 2.8588        & 0.9936                     & 0.0082         \\
31  bands         & 6 bands           & 47.27           & 0.9912          & 1.8545          & 2.799         & 0.9945                     & 0.0081         \\
61  bands         & 6 bands           & 47.31           & 0.9912          & 1.8261          & 2.8198        & 0.9946                     & 0.0081         \\
121  bands        & 6 bands           & 47.24           & 0.991           & 1.8519          & 2.8949        & 0.9944                     & 0.0084         \\
61  bands         & 4 bands           & 45.56           & 0.986           & 2.0299          & 3.1919        & 0.9914                     & 0.0092         \\
121 bands         & 4 bands           & 45.62           & 0.9859          & 2.0476          & 3.2506        & 0.9913                     & 0.0094         \\
31 bands          & 4 bands           & 45.47           & 0.9859          & 2.0517          & 3.1815        & 0.9913                     & 0.0092         \\
16 bands          & 4 bands           & 45.21           & 0.9858          & 2.0456          & 3.1974        & 0.9913                     & 0.0092         \\
121 bands         & 3 bands           & 42.54           & 0.9768          & 4.2612          & 3.6212        & 0.975                      & 0.0109         \\
61  bands         & 3 bands           & 42.51           & 0.9771          & 4.2225          & 3.5594        & 0.9755                     & 0.0107         \\
31  bands         & 3 bands           & 42.47           & 0.977           & 4.2165          & 3.5467        & 0.9755                     & 0.0107         \\
16  bands         & 3 bands           & 42.37           & 0.9768          & 4.2571          & 3.5744        & 0.9751                     & 0.0108         \\ \hline\hline
\end{tabular}

\caption{Reconstruction encoder  → decoder results on different numbers of RWA compacted bands and PCA compacted bands.}
\end{table*}
\subsection{Structure of U-Net in GEWDiff framework.} In previous experiments, we found that the simple model did not have enough ability to preserve spectral information. 3D U-Net performs convolutions in both spatial and spectral dimensions. Thus, it can capture joint spatial-spectral features from PCA components in latent space, which is critical for accurately modeling hyperspectral images.
\begin{figure*}[]
  \centering
  \includegraphics[width = 0.7\linewidth]{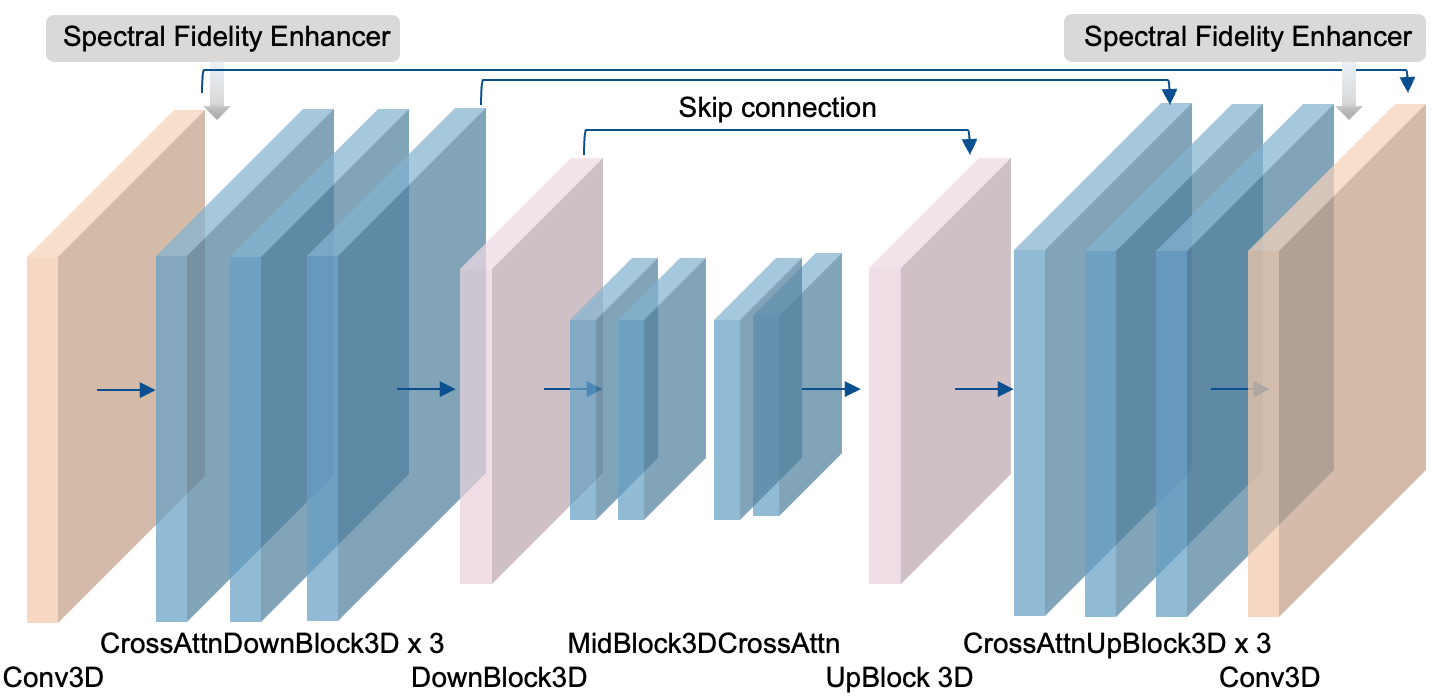}
  \caption{Illustration of three-dimensional U-Net structure.}
  \label{fig:unet}
\end{figure*}

Additionally, we introduce a Spectral Fidelity Enhancer (SFE) module into the diffusion model to further preserve the spectral structure of hyperspectral images. The SFE can be intuitively understood as a spectral attention mechanism that emphasizes key spectral features during the generation process. By guiding the model to focus on critical wavelength information, the SFE helps reduce spectral distortion and enhances the fidelity of the reconstructed spectral profiles.

\section{Hyperparameter analysis in sampling stage}
In this section, we give a more detailed explanation of the hyperparameters that were used in our method.
\subsection{Effect of {$\rho$} on sampling and spectral fidelity.} 
\begin{figure}[!ht]
  \centering
  \includegraphics[width = 1.0\linewidth]{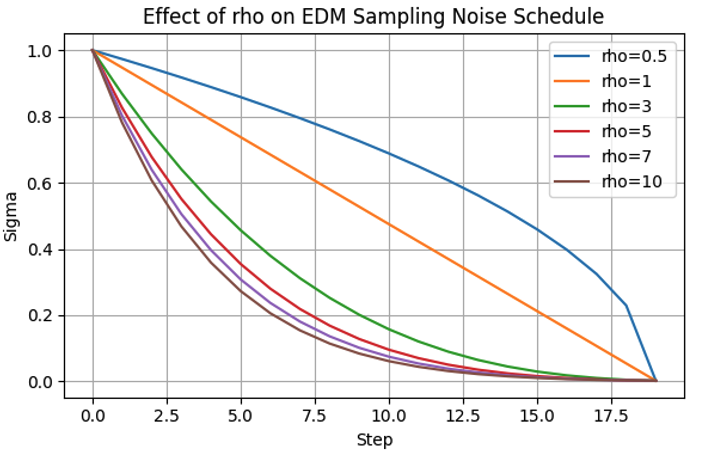}
  \caption{Effect of $\rho$ on EDM sampling noise schedule.}
  \label{fig:rho}
\end{figure}

In EDM, the parameter $\rho$ governs the curvature of the noise schedule that interpolates between the maximum noise level $\sigma_{\text{max}}$ and the minimum noise level $\sigma_{\text{min}}$ across sampling steps. 
The $\rho$ of EDM does not decrease linearly from $\sigma_{\text{max}}$ → $\sigma_{\text{min}}$, but is calculated by the following formula:
\begin{equation}
    \sigma _{n}=\left ( \sigma _{\textrm{max}}^{1/\rho } + \frac{n}{N-1}\left ( \sigma _{\textrm{min}}^{1/\rho }-\sigma _{\textrm{max}}^{1/\rho } \right )\right )^{\rho }, 
\end{equation}
Specifically, $\rho$ determines whether these $\sigma$ decrease linearly ($\rho=1$) or nonlinearly ($\rho>1$), as shown in Figure. \ref{fig:rho}

$\rho$ controls how the noise levels decay, with larger values resulting in a schedule that rapidly decreases at early steps and flattens at later steps. This structure allows more refinement in the low-noise regime, which typically enhances perceptual detail and image fidelity. However, in the context of conditional hyperspectral image synthesis, maintaining spectral consistency across all bands is crucial. High values of $\rho$ tend to introduce increased stochasticity in the denoising process, especially in high-frequency components, potentially leading to unwanted spectral artifacts or incoherent band-wise variations. We designed a group of experiments on the validation dataset. To mitigate this, we adopt a relatively low value $\rho\in [0.6,0.7]$, which produces a more linear and uniform noise schedule. This setting constrains the denoising trajectory, suppressing random high-frequency noise while promoting better alignment with the spectral distribution of the conditioning data. As a result, it helps preserve spectral fidelity and structural coherence in the reconstructed hyperspectral images.
\begin{figure*}[!ht]
  \centering
  \includegraphics[width = 0.8\linewidth]{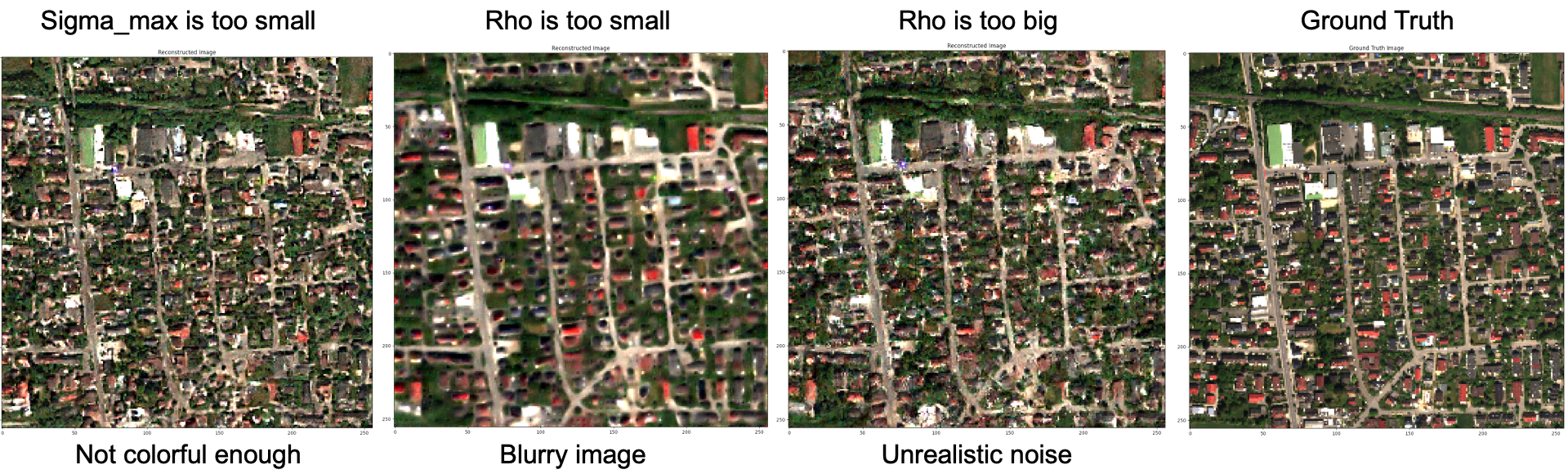}
  \caption{Three typical effects of unsuitable parameters on validation set (informative only).}
  \label{fig:bad}
\end{figure*}
\subsection{Effect of $\sigma_{\textrm{max}}$ and $\sigma_{\textrm{min}}$ on sampling and spectral fidelity.} $\sigma_{\textrm{max}}$ and $\sigma_{\textrm{min}}$ are the two parameters that control the maximum noise strength and minimum noise strength in the sampling stage. At this point, the diffusion model has learned the ability to denoise within a specific noise scale range. We test the effect of these two parameters on the sampling stage. $\sigma_{\textrm{max}}$ represents the initial loss, which will affect the ``divergence" of the image. A higher $\sigma_{\textrm{max}}$ causes higher initial noise, higher image freedom. The result image may contain more details, but with large amounts of noise. A lower $\sigma_{\textrm{max}}$ causes more conservative results, and may lose color information. We adopted $\sigma_{\textrm{max}}=80$ as the final setting. $\sigma_{\textrm{min}}$ represents the end noise, which will affect the ``convergence" of the image. The $\sigma_{\textrm{min}}$ is large, the convergence is shorter, and the image is more blurry. Smaller $\sigma_{\textrm{min}}$ causes longer convergence process. Reconstruction images contain more details but also have more artifacts. The best performance $\sigma_{\textrm{min}}$ range is [0.002, 0.2].

\subsection{Training and testing settings.}Our training and testing pipeline involves several key hyperparameters that influence both performance and computational efficiency. For spectral dimensionality reduction, we first compact the hyperspectral data using a wavelet transform, retaining 121 bands (compact bands), followed by a PCA transformation during training to further reduce the spectral dimension to 20 bands (PCA bands). Training is conducted with a batch size of 1 and typically runs for 200 epochs. The number of GPUs (num processes) can vary from 1 to 4, depending on available hardware. To improve learning quality, we apply a combination of loss functions: pixel loss (l1 lambda = 0.8), perceptual loss (l2 lambda = 0.1), and gradient loss (l3 lambda = 0.1). We also include optional modules such as mask conditioning (mask) and edge perturbation (edge) to enhance robustness. During inference, the sampling process is controlled by the number of diffusion steps (timesteps), typically set to 50 for a balance between quality and speed. Noise scheduling is governed by $\sigma_{\textrm{min}}$ (0.002–0.2), $\sigma_{\textrm{max}}$ (80–90), and $\sigma_{\textrm{data}}$ (0.5), while the parameter rho (0.6–0.7) adjusts the curvature of the sampling trajectory. Additionally, the recall option allows resuming training from any specific epoch. These parameters are tuned to achieve a balance between computational cost and reconstruction fidelity.

Table 3 and Figure \ref{fig:bad} present a summary of hyperparameter effects and visualize the outcomes of suboptimal parameter settings. Before the camera-ready version is finalized, we will release the model checkpoints to support reproducibility. We encourage users to adjust these hyperparameters for their specific datasets, eliminating the need for retraining.

\setlength{\tabcolsep}{2mm}
\begin{table*}[]
\centering
\label{tab:para}
\fontsize{9pt}{9pt}\selectfont
\begin{tabular}{cccc}
\hline \hline
Parameter  & Higher                                                                          & Lower                                                                            & Influence                                         \\ \hline
$\sigma_{\textrm{max}}$ & Larger initial noise, high image freedom,       & The initial noise is small, more conservative, & This will affect the \\
&more details, but with larger noise. & and may lose color information.&  "divergence" of the image.\\
$\sigma_{\textrm{min}}$ & The end noise is large, the convergence       & Smaller ending noise and cleaner                                     & Smaller is cleaner, but also           \\
&is poor, and the image is blurry.&convergence.&more risky.\\
$\rho$       & The number of steps is denser in the early  & The number of steps is sparse in the early    & Control the curvature of the  \\
&stage and sparser in the late stage.& stage and denser in the late stage.&  "sampling curve". \\
$t$   & Larger and more precise, but slower                                 & The sampling speed is fast,                      & Affects the total number of   \\ 
& to compute.&but it is easy to be blurry.  &steps for denoising.\\ \hline\hline
\end{tabular}
\caption{Summary of hyperparameter effects (informative only)}
\end{table*}

\subsection{Robustness under imperfect conditioning.}
To verify the robustness of GEWDiff under imperfect conditions, we conducted controlled perturbation experiments on mask input and low-resolution (LR) images to simulate real segmentation and sensor noise. Specifically, we applied two methods: (1) random mask erosion/dilation (1–3 pixels) combined with random spatial translation (±1–2 pixels) to assess its sensitivity to geometric boundary errors; and (2) introducing 1\% additive Gaussian noise into the low-resolution input to test its spectral stability. Quantitative evaluation results in Table \ref{tab:p} show that segmentation perturbation leads to a 0.03 dB decrease in PSNR and an increase in FID by 11 (MDAS1), while noisy low-resolution input leads to a 0.2 dB decrease in PSNR and an increase in FID by 1.5. Notably, the structural fidelity remains essentially unchanged under these perturbations, indicating that GEWDiff's cross-scale geometric prior and diffusion denoising process effectively mitigates moderate mask errors and input noise. Although extreme segmentation failures may still affect the reconstruction results, these results demonstrate that GEWDiff can generalize beyond ideal mask conditions and maintain stable performance under real noise and boundary deviations. Moreover, the mask and NDVI used in our work will not generate extra data demand, as they were derived from low-resolution images. In future work, we will explore uncertainty-aware mask conditionalization and adaptive attention weighting to further improve their robustness to highly ambiguous or noisy inputs.
\setlength{\tabcolsep}{1mm}
\begin{table}[]
\fontsize{9pt}{9pt}\selectfont
\begin{tabular}{cccccc}
\hline\hline
\textbf{Dataset}        & \textbf{Methods} & \multicolumn{2}{c}{\textbf{Mask perturbation}} & \multicolumn{2}{c}{\textbf{Noise perturbation}} \\ \hline
\multirow{7}{*}{MDAS 1} & PSNR↑            & 28.8372                & 0.0544                & 28.6                   & 0.12                   \\
                        & SSIM↑            & 0.7286                 & 0.0032                & 0.719                  & 0.005                  \\
                        & SAM ↓            & 8.5045                 & 0.0323                & 8.684                  & 0.008                  \\
                        & CC↑              & 0.8113                 & 0.0025                & 0.803                  & 0.005                  \\
                        & RMSE ↓           & 0.05246                & 0.00028               & 0.05313                & 0.00067                \\
                        & FID ↓            & 56.8863                & 0.8352                & 46.05                  & 5.3                    \\
                        & LV ↑             & 0.003698               & 0.000022              & 0.0037                 & 0.00009                \\ \hline\hline
\end{tabular}
\caption{Conditions perturbation experiments}
\label{tab:p}
\end{table}
\subsection{Practical utility on downstream land-cover classification task}
\begin{figure}[!ht]
  \centering
  \includegraphics[width = 1\linewidth]{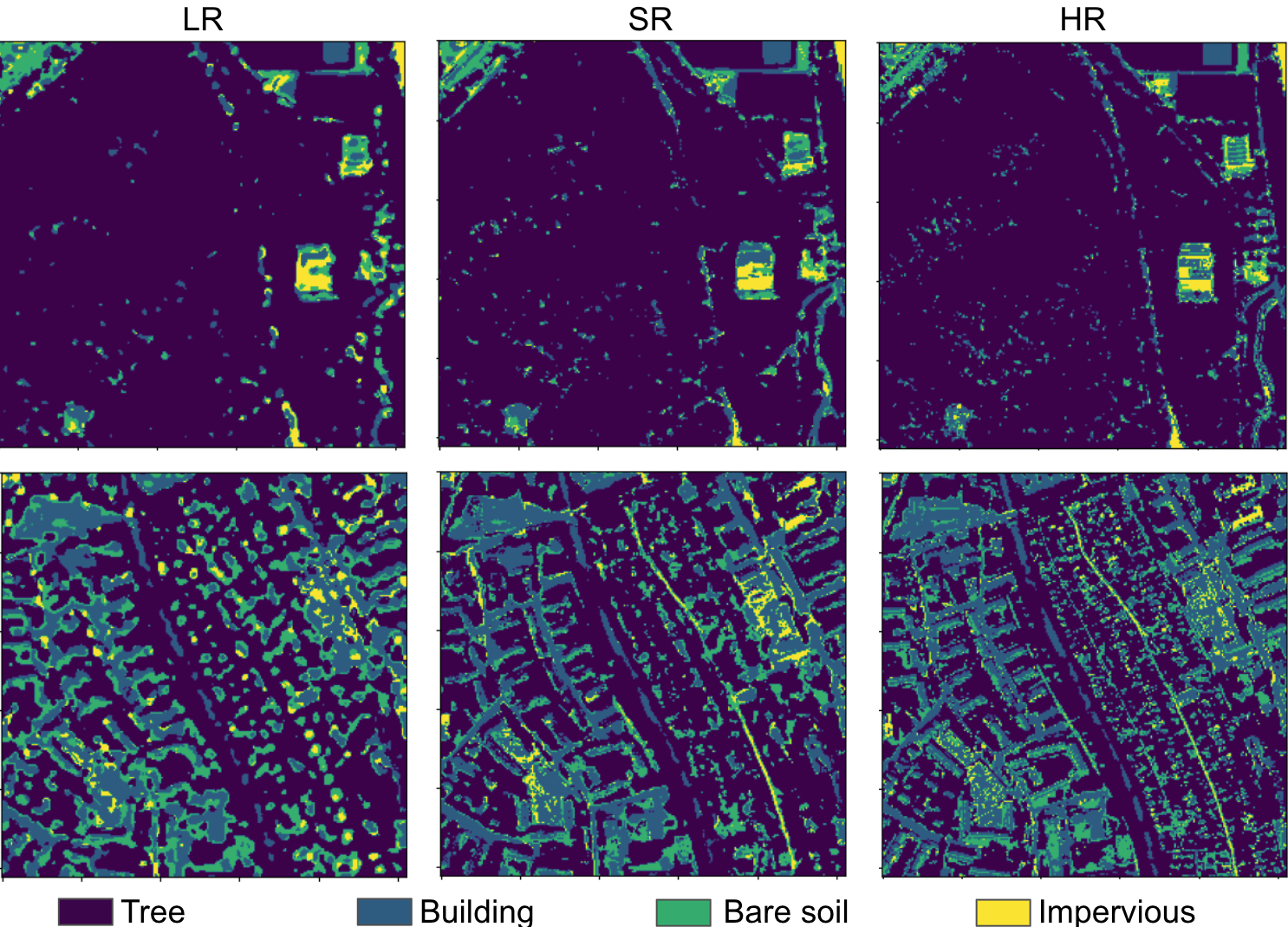}
  \caption{Classification map of low-resolution image, super-resolution image, and high-resolution image.}
  \label{fig:c}
\end{figure}

To further evaluate the practicality of GEWDiff, we assessed its impact on downstream hyperspectral land cover classification tasks. We randomly sampled 30 points covering four semantic categories (trees, buildings, bare soil, impermeable surfaces) and trained a random forest classifier using super-resolution output (SR) as input. For comparison, we applied the same classifier configuration and sampling scheme to the original 10-meter low-resolution (LR) input. Experiments were conducted on two independent samples (MDAS 1 and MDAS 2), and the classification results were benchmarked against pseudo-ground values extracted from high-resolution data, as shown in Figure \ref{fig:c}.

In both samples, GEWDiff consistently improved the accuracy of semantic classification. On MDAS1, SR improved the average IoU from 0.4312 to 0.4486 and the average F1 score from 0.5553 to 0.5761, achieving more accurate class boundary delimitation while maintaining the same overall accuracy (OA) as the LR baseline (0.9018). On the more challenging MDAS2 samples, SR significantly improved OA (0.6996 → 0.7222), mIoU (0.3967 → 0.4181), and mF1 (0.5319 → 0.5572). Visual comparison (Appendix Figure X) reveals a clearer distinction between built-up areas and vegetation, as well as a more refined detection of impermeable structures. These results demonstrate that GEWDiff does not merely upsample spectral data, but generates representations capable of distinguishing more class features and improving downstream discriminative capabilities. Therefore, even with limited supervision information and a small number of training samples, this model can bring tangible benefits to practical remote sensing applications.

\end{document}